\documentclass{article} 
\usepackage{iclr2027_conference,times}

\usepackage{amsmath,amsfonts,bm}

\def\eqref#1{equation~\ref{#1}}

\def\1{\bm{1}}

\DeclareMathAlphabet{\mathsfit}{\encodingdefault}{\sfdefault}{m}{sl}
\SetMathAlphabet{\mathsfit}{bold}{\encodingdefault}{\sfdefault}{bx}{n}

\usepackage{hyperref}
\usepackage{url}
\usepackage{algorithm}
\usepackage{adjustbox}
\usepackage{algpseudocode}
\usepackage{amssymb}
\usepackage{booktabs}
\usepackage{graphicx}
\usepackage{multirow}
\usepackage{xcolor}
\usepackage{enumitem}
\usepackage{placeins}
\usepackage{wrapfig}

\usepackage{soul}

\newcommand{\modelname}{ReferClaw}

\newcommand{\eg}{\textit{e.g.}}
\newcommand{\ie}{\textit{i.e.}}
\newcommand{\vers}{\textit{vs. }}

\newcommand{\cmmnt}[1]{}

\title{ReferClaw: Disambiguating User Intent for Egocentric Assistance}

\author{
    Hangyu Qin$^1$, Junbin Xiao$^2$\footnotemark[1], Angela Yao$^1$ \\
    $^1$National University of Singapore $^2$University of Science and Technology of China\\
    \\
}

\iclrfinalcopy 
\begin{document}
\footnotetext[1]{Corresponding author: 
\texttt{junbinxiao@ustc.edu.cn}}

\maketitle

\begin{abstract}
Users of egocentric AI assistants often ask short, ambiguous questions whose intended referents cannot be uniquely resolved from video and dialogue history. For example, “Where did I leave it?” may refer to several previously handled objects. Existing VideoQA systems typically commit to a single interpretation, risking plausible but intention-misaligned answers. We introduce \modelname, an agentic framework that formulates egocentric streaming QA as an interactive process: the assistant asks clarifying questions to resolve user intent before answering. Three cooperating agents identify candidate interpretations, request clarification, and generate answers grounded in the resolved intent and visual evidence. A bounded, cost-aware policy balances expected information gain against user effort while limiting clarification turns and question complexity. \modelname~learns from interaction through intent-specific skill updates and process-reward-guided policy optimization with agent-level credit assignment. Evaluations on augmented egocentric QA benchmarks demonstrate improved answer accuracy and joint answer-and-grounding performance over existing baselines, alongside zero-shot transfer across datasets. These findings support interactive clarification as a practical approach to resolving online user intent in egocentric streaming assistance.
\end{abstract}
\section{Introduction}
\label{sec:introduction}

\begin{figure}[t!]
    \centering
    \includegraphics[width=\linewidth]{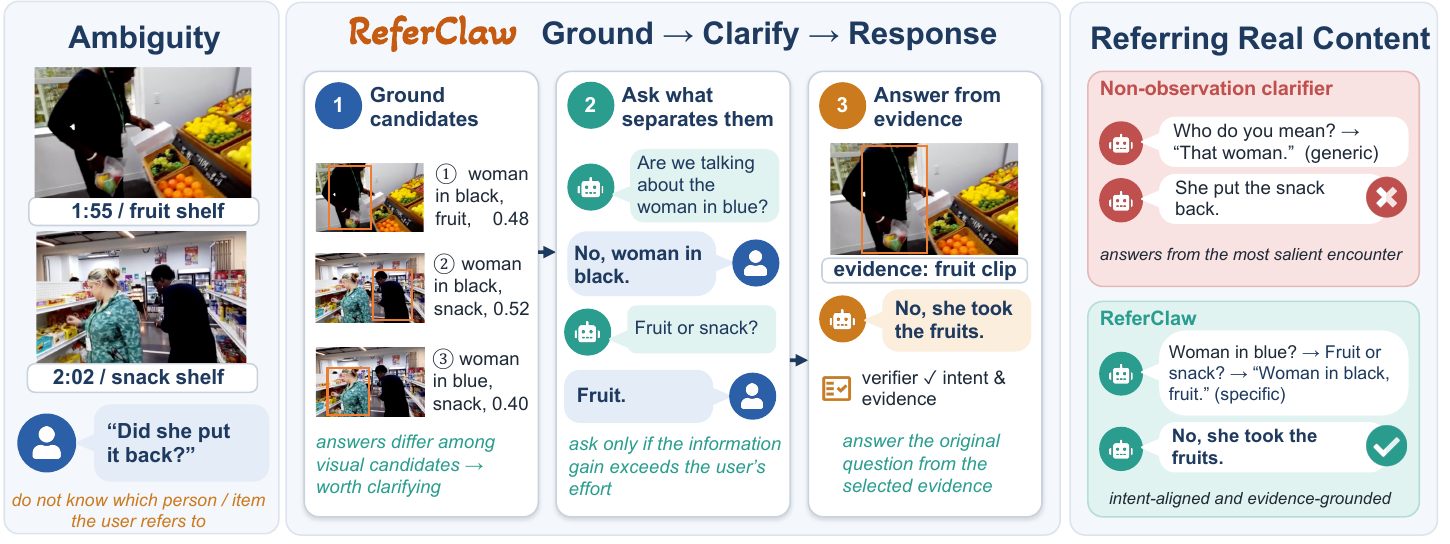}
    \vspace{-0.4cm}
    \caption{Different encounters can support different interpretations of the same question. \modelname~asks for clarification to identify the intended encounter and select evidence for answering.}
    \vspace{-0.4cm}
    \label{fig:teaser}
\end{figure}

Egocentric cameras paired with multimodal AI assistants are becoming natural interfaces to the physical world, enabling continuous visual memory, navigation, and daily assistance.
Recent benchmarks have advanced episodic memory~\citep{barmann2022did,yang2025egolife}, long-range understanding~\citep{mangalam2023egoschema}, and situated assistance~\citep{xiao2025egoblind,sun2025visual,qin2026buddyvqa}. Their standard QA protocols, however, evaluate answers against fixed questions and observation histories, rather than how an assistant acquires missing intent information. In practice, user requests are brief and ego-deictic, and the available context may support several interpretations. For ``\texttt{Did she put it back?}'' (Fig.~\ref{fig:teaser}), different persons may perform the same action at different times. Even when both encounters are visible, the intended one may remain unclear. An answer can therefore be compatible with the video yet irrelevant to what the user meant.

Despite this progress, the challenge remains largely underexplored. Existing approaches \citep{peng2025eye,choi2026you,qin2026buddyvqa} exploit eye gaze and hand gesture as visual or textual prompts for disambiguating user intent. While effective, the gains remain limited for the complexity of real-world interactions. Recent VLMs instead explore proactive request for clarification ~\citep{jian2025clearvqa,cao2026clarifyanswer,yang2026clarifier}, showing that models can ask rather than guess based on the utterance and a single observation.  This may suffice for a static image, but may not for dynamic video stream, where different encounters can support different interpretations of the same question (Fig.~\ref{fig:teaser}). We therefore study how explicit spatio-temporal interpretations can guide what to ask, how to interpret the response, and which evidence to use for answering.

We formulate intent disambiguation as bounded information acquisition over grounded candidate interpretations. 
We propose \modelname\ which coordinates three agents around these candidates. The Intention Grounding Agent (IGA) constructs the candidate intents by grounding the question against the observed video, pairing each candidate with its supporting visual evidence, and scores the candidates by evidential support; the Cost-Aware Clarification Agent (CCA) generates and chooses the best clarification question by balancing expected information gain against respondent effort; and the Evidence-Fused Answering Agent (EFA) answers the original question from the resolved intent and its evidence. Their shared state allows clarification responses to refine the candidates or trigger re-grounding. A deterministic controller bounds clarification turns and question complexity, and decides when to ask or answer.
\modelname~also learns from past interactions to reduce recurring disambiguation errors. Following MetaClaw's two-timescale design~\citep{xia2026metaclaw}, verified feedback produces intent-specific skills that take effect immediately and process rewards that guide updates to the responsible agent. The skills encode which contrasts distinguish competing interpretations, while agent-level updates target failures in grounding, clarification, or answering. In our evaluation protocol, the annotated target intent, supporting evidence, and answer correctness become available only after each prediction, so updates affect subsequent interactions.

To facilitate the study, we augment two recent egocentric streaming QA benchmarks BuddyVQA ~\citep{qin2026buddyvqa} and EgoBlind~\citep{xiao2025egoblind} with ambiguity labels, target intents, and spatio-temporal visual evidence. Experiments separate the benefits of clarification from learning, test ambiguity across repeated encounters, and assess transfer to independent image-based clarification benchmarks ClearVQA~\citep{jian2025clearvqa} and ContextClarify~\citep{cao2026clarifyanswer}. We additionally evaluate well-specified questions and human respondents to examine unnecessary interruptions and the applicability of simulated interactions.
Our contributions are threefold:
\begin{itemize}[leftmargin=*]
    \item We formulate intent disambiguation in egocentric QA assistance as bounded information acquisition over spatio-temporally grounded interpretations.
    \item We develop \modelname, which coordinates grounding, clarification, and answering agents through shared candidates, and learns from feedback through intent-specific skills and agent-level updates.
    \item We augment two ego-streaming QA benchmarks with intent and grounding annotations, and evaluate clarification, continual adaptation, and transfer under controlled interaction protocols.
\end{itemize}

\section{Related Work}
\label{sec:related-work}

\noindent\textbf{Clarification and User Intent Disambiguation.}
Clarification establishes common ground by acquiring missing information and resolving alternative interpretations~\citep{clark1991grounding,saeidi2018sharc,xu2019claqua,rao2018evpi,min2020ambigqa,lee2023clarification}. Cooperative visual dialog agents~\citep{das2017cooperative} and GuessWhat~\citep{devries2017guesswhat} extend goal-directed question asking to images, and later methods select questions through uncertainty~\citep{abbasnejad2019whats} or expected information gain~\citep{white2021clarification}. Recent work localizes plausible image regions for ambiguous questions~\citep{chen2025focusambiguity}, teaches VLMs to request clarification on ClearVQA~\citep{jian2025clearvqa}, and learns ask-or-answer policies on ContextClarify~\citep{cao2026clarifyanswer}. Plug-and-Play Clarifier~\citep{yang2026clarifier} addresses linguistic ambiguity, capture quality, and pointing gestures in egocentric interaction. Existing clarification benchmarks are largely image-based, motivating our augmentation of BuddyVQA and EgoBlind with intent annotations for video-based disambiguation (Sec.~\ref{sec:evaluation-benchmarks}). Methodologically, prior works clarify 
from the utterance and a single observation; \modelname~instead focuses on competing interpretations grounded across video history, using shared candidates to connect clarification with evidence selection under bounded respondent effort.

\noindent\textbf{Egocentric and Grounded Video QA.}
Ego4D~\citep{grauman2022ego4d} evaluates episodic memory, while EgoSchema~\citep{mangalam2023egoschema} tests long-range egocentric video understanding. EgoBlind~\citep{xiao2025egoblind} emphasizes situated assistance, and BuddyVQA~\citep{qin2026buddyvqa} evaluates ego-deictic and history-dependent questions over a causal stream. GroundVQA~\citep{di2024groundvqa} and NExT-GQA~\citep{xiao2024nextgqa} couple answering with temporal localization, GeLM~\citep{chen2025gelm} develops grounded multi-hop reasoning, D$^2$VLM~\citep{zeng2025d2vlm} factorizes grounding and answering, and recent systems expose explicit temporal or spatio-temporal evidence for verifiable answers~\citep{meng2025openo3video,seo2026groundformer,huang2026egvqa}. Related systems support interactive video questioning~\citep{wang2025rga3}, pixel-level conversational grounding~\citep{munasinghe2025videoglamm}, and streaming evidence retrieval~\citep{di2025rekv}. These methods localize evidence under a single interpretation of the query, so stronger retrieval may confidently ground the wrong referent when several intentions remain plausible. \modelname~represents the alternatives explicitly and acquires 
clarifications on which interpretation to guide evidence selection and answering.

\noindent\textbf{Multi-Agent Collaboration. } 
Multi-agent systems coordinate specialized roles through structured communication~\citep{hong2024metagpt,chen2024agentverse,liu2025videomind,dang2025mupa}. Interaction-based adaptation uses self-feedback, verbal reflection, and reusable skills or routines~\citep{madaan2023selfrefine,shinn2023reflexion,zhao2024expel,wang2024voyager,wang2025awm,tu2026visualclaw}. MetaClaw combines fast skill adaptation with slow policy optimization~\citep{xia2026metaclaw}, while process supervision rewards intermediate steps~\citep{lightman2024verify,zhang2025prm}. \modelname~adapts these mechanisms to intent disambiguation: skills encode ambiguity triggers and decisive contrasts, while process rewards guide updates to the grounding, clarification, and answering agents.

\section{Method} \label{sec:method}

Clarifying an ambiguous question requires three decisions that ordinary VideoQA does not: whether the question admits more than one visually supported interpretation, what to ask so that the user can resolve it at little effort, and how to answer from the interpretation the user selects. \modelname\ assigns each decision to a dedicated agent operating on shared grounded candidate interpretations, coordinated by a fixed controller that bounds the user's burden (Fig.~\ref{fig:overview}), and learns from verified interaction feedback so that disambiguation mistakes are not repeated. 

\subsection{Problem Setup}\label{sec:problem-setup}
For a question $q$ asked at video time $s_q$, the assistant observes the causal prefix $V_{\le s_q}=\{v_s\}_{s=1}^{s_q}$ and episode dialogue $D$. The prefix stays fixed during clarification. At turn $t$, $h_t=\{(c_\ell,y_\ell)\}_{\ell=0}^{t-1}$ records clarification questions $c$ and responses $y$. Intent is ambiguous when multiple visually supported interpretations remain plausible given $(V_{\le s_q},q,D,h_t)$.

\begin{figure}[t]
    \centering
    \includegraphics[width=1.0\linewidth]{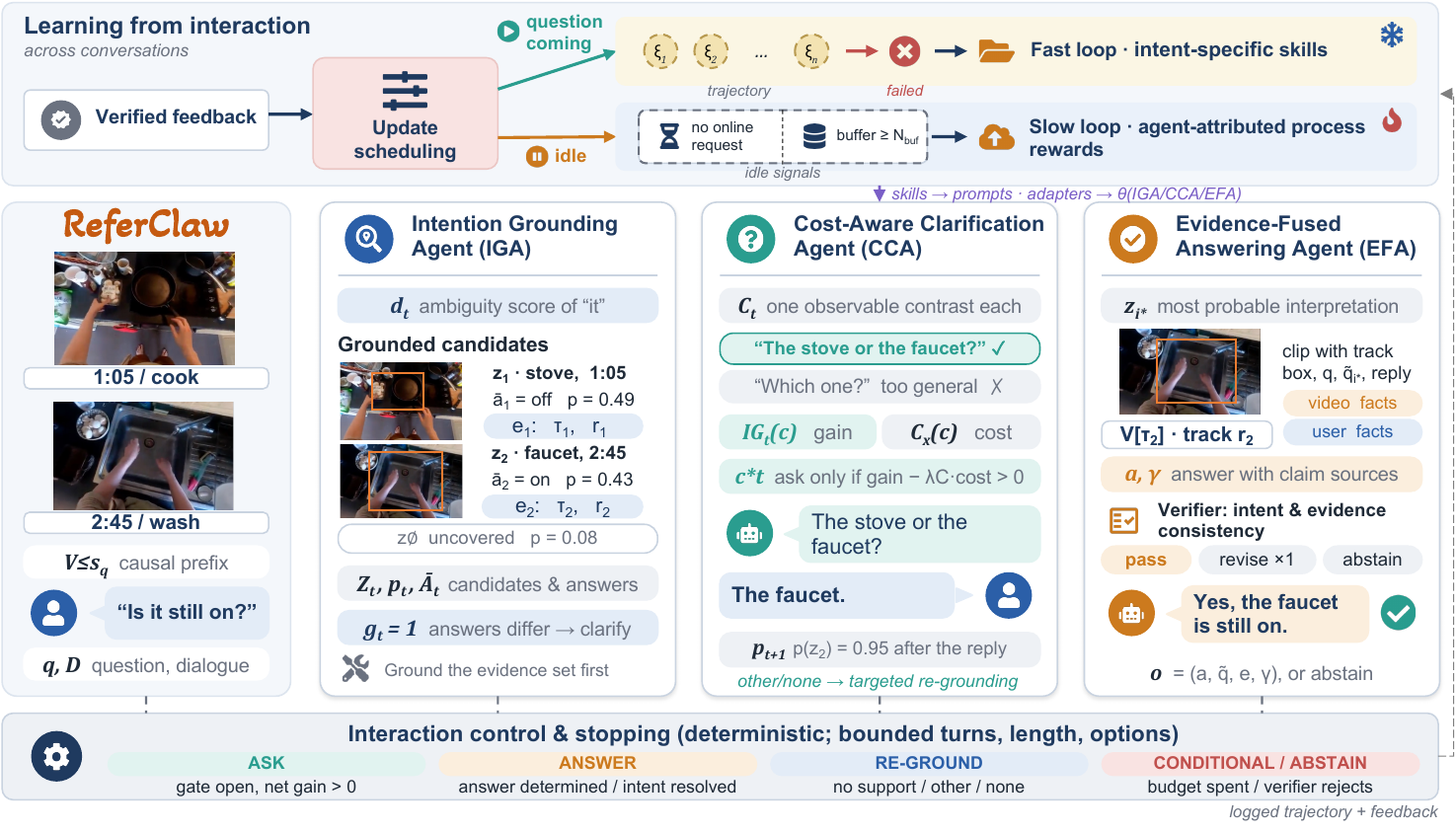}
    \vspace{-0.6cm}
    \caption{Overview of \modelname. The system coordinates three agents (IGA, CCA, EFA) to respectively ground candidate interpretations, requests clarification when useful, and verifies answers against the resolved intent and evidence. A fixed controller bounds interaction. \emph{Top:} verified feedback updates prompt-level skills immediately and agent-specific LoRA adapters during idle periods.}
    \label{fig:overview}
\end{figure}

A candidate $z_i=(\widetilde q_i,e_i)$ pairs a question rewrite $\widetilde q_i$ with evidence $e_i=(\tau_i,r_i)$: an interval $\tau_i\subseteq[1,s_q]$ and referent box track $r_i$ (empty for non-localizable information). For ``Is it still on?'' in Fig.~\ref{fig:overview}, rewrites may specify the ``stove'' used at time 1:05 or the ``faucet'' at 2:45. The assistant maintains $\mathcal Z_t=\{z_i\}_{i=1}^{n_t}$, $n_t\le N_{\max}$, and belief $p_t$ over these candidates plus an uncovered interpretation $z_\varnothing$. It asks $c_t$ or terminates with an \textsc{Answer} $(a,\widetilde q,e,\gamma)$, where $\gamma$ is the belief mass supporting the answer, a \textsc{Conditional} output with at most two interpretations and answers, or \textsc{Abstain}.

For policy $\pi$ and respondent type $x\in\{\mathrm H,\mathrm L\}$ (human or LLM), we seek
\begin{equation}
\max_{\pi}\;
\mathbb E\big[U_{\mathrm{ans}}
+\lambda_g\, U_{\mathrm{grd}}
-\textstyle\sum_{t=0}^{N_{\mathrm{ask}}-1} C_x(c_t)\big]
\quad\text{s.t.}\quad
N_{\mathrm{ask}}\le T_{\max},\; |c_t|\le L_{\max},\; m(c_t)\le M_{\max},
\label{eq:referclaw-objective}
\end{equation}
Here $U_{\mathrm{ans}}(a,z^\star)$ is binary judged correctness for target intent $z^\star$, and $U_{\mathrm{grd}}(e,e^\star)=\min(\mathrm{tIoU},\mathrm{sIoU})$ measures overlap with target evidence $e^\star$, using temporal IoU alone when no target box exists. The weight $\lambda_g$ balances grounding; $C_x(c_t)\in[0,1]$ measures respondent cost. \textsc{Conditional} and \textsc{Abstain} receive zero utility. The limits bound the number of questions $N_{\mathrm{ask}}$, question length $|c_t|$ in tokens, and displayed options $m(c_t)$, preventing endless clarification and one-shot ``super-questions''  that overwhelm the user with many alternatives in a single turn.

\subsection{Cooperating Agents and Interaction Controller}\label{sec:agents}
The three agents 
share $(\mathcal Z_t,p_t)$. Each uses a role-specific prompt and LoRA adapter on one frozen multimodal backbone; a fixed controller sequences their actions and enforces the interaction budget.

\textbf{Intention Grounding Agent.}\label{sec:iga}
IGA predicts an ambiguity score $d_t\in[0,1]$ and responsible span, then grounds before rewriting. Temporal retrieval, detection, and tracking produce up to eight evidence cards containing an interval--track pair, grounding score $\rho_i$, and visible-content description (Appendix~\ref{app:candidates}). Each card yields a rewrite $\widetilde q_i$, support score $s_i\in[0,1]$, and provisional answer $\bar a_i$. After merging duplicate entity--event cards and retaining at most $N_{\max}$ candidates, IGA initializes
\begin{equation}
    p_t(z_i)=(1-\epsilon_t)\,\frac{\exp(s_i/\kappa)}{\sum_{j=1}^{n_t}\exp(s_j/\kappa)},
    \qquad
    p_t(z_\varnothing)=\epsilon_t=\epsilon_0\,\bigl(1-\max\nolimits_i \rho_i\bigr),
    \label{eq:intent-posterior}
\end{equation}
where $\kappa$ is a validation-fitted temperature and $\epsilon_t$ reserves uncovered-intent mass when grounding support is weak. Clarification requires high ambiguity, no dominant candidate, and disagreement among provisional answers:
\begin{equation}
g_t=\mathbb I[d_t\ge\tau_d]\;
    \mathbb I[1-\max\nolimits_i p_t(z_i)\ge\tau_u]\;
    \mathbb I[\operatorname{Dis}(\overline{\mathcal A}_t)\ge\tau_a],
\label{eq:clarification-gate}
\end{equation}
Here $\operatorname{Dis}$ is the belief-weighted fraction of candidate pairs with different provisional answers $\overline{\mathcal A}_t=\{\bar a_i\}$. When all plausible candidates imply the same answer, clarification is unnecessary even if the intended referent remains uncertain.

\textbf{Cost-Aware Clarification Agent.}\label{sec:cca}
CCA proposes at most six questions $\mathcal C_t$, each contrasting candidates by identity, occurrence, location, or action. Each declares an option-based partition with an ``other/none'' option for $z_\varnothing$. The response model $p_x(y\mid c,z_i,h_t)$ predicts the respondent's answer class: a partition cell, other/none, uncertain, or invalid/missing. For LLM respondents, smoothed frequencies from eight backbone samples estimate this likelihood, conditioning on candidate $z_i$ rather than on the annotated target intent, which only the respondent supplying actual responses (simulator or human) receives.

An actual response $y_t$ updates the belief by Bayes' rule. Hypothetical responses define expected information gain (IG), measured as the reduction in Shannon entropy $H$ (bits):
\begin{equation}
    p_{t+1}(z_i)\propto p_t(z_i)\,p_x(y_t\mid c_t,z_i,h_t),
    \quad
    \operatorname{IG}_t(c)=H(p_t)-\textstyle\sum_{y} p_x(y\mid c,h_t)\,H\bigl(p_{t+1}(\cdot\mid c,y)\bigr),
    \label{eq:posterior-update}
\end{equation}
where $p_x(y\mid c,h_t)=\sum_i p_t(z_i)p_x(y\mid c,z_i,h_t)$. IG is a one-step surrogate for answer utility, used only when the gate detects answer disagreement. CCA balances it against normalized effort $\overline E_x$, delay $\overline T_x$, and failure risk $\overline F_x$:
\begin{equation}
    C_x(c)=w_E\,\overline E_x(c)+w_T\,\overline T_x(c)+w_F\,\overline F_x(c),
    \qquad
    c_t^\star=\arg\max_{c\in\mathcal C_t}u_t(c),
    \label{eq:simple-cost}
\end{equation}
where $u_t(c)=\operatorname{IG}_t(c)-\lambda_C\,C_x(c)$ is the net utility of a question. CCA returns $c_t^\star$ together with $u_t(c_t^\star)$, which the controller uses to decide whether to ask, or \emph{no-question} when the pool is empty or $u_t(c_t^\star)\le0$.
Human response likelihoods and costs are calibrated on the user study using question length, choice complexity~\citep{hick1952,hyman1953}, response time, and failure rate. LLM costs use token counts, serving latency, and predicted invalid-format mass (Appendices~\ref{app:cost-details} and~\ref{app:human-calibration}).

\textbf{Evidence-Fused Answering Agent.}\label{sec:efa}
EFA selects $i^\star=\arg\max_i p_t(z_i)$ and generates $a=F_\omega(V[\tau_{i^\star}],r_{i^\star},q,\widetilde q_{i^\star},h_t)$ with confidence $\gamma=p_t(z_{i^\star})$. Its prompt separates observations from user responses and labels claims as \texttt{video} or \texttt{user}: clarification identifies the appliance, not whether it is on. A separate verifier prompt checks alignment with the rewrite and support for each claim in its corresponding source: \texttt{video} claims must be visible in the evidence card, and \texttt{user} claims must appear in $h_t$. It permits one revision before passing or abstaining, without changing the interpretation or retrieving evidence (Appendices~\ref{app:efa-details} and~\ref{app:verifier}).

\textbf{Interaction Controller.}
In priority order, the controller abstains if support remains inadequate after one re-grounding; invokes EFA if $g_t=0$ or $\max_i p_t(z_i)\ge\tau_p$; and asks if $g_t=1$, $t<T_{\max}$, and net utility is positive. Otherwise, it returns \textsc{Conditional} when the top two candidates disagree and their belief margin is below $\tau_g$, or invokes EFA. Responses update Eq.~\ref{eq:posterior-update}; other/none or an unsupported constraint triggers one constrained re-grounding. Uncertain or invalid responses leave belief unchanged and ban the asked partition (Appendix~\ref{app:controller-alg}).

\subsection{Learning From Interaction}\label{sec:learning}

Following MetaClaw~\citep{xia2026metaclaw}, our learning maintains a versioned skill library $\mathcal S_g$ and agent adapters $\theta_g$ indexed by skill generation $g$. Under our prequential protocol (Sec.~\ref{sec:baselines-implementation}), $z^\star$, $e^\star$, and the judge's verdict are revealed only \emph{after} a prediction is logged. A trajectory whose answer is wrong, whose evidence misses $e^\star$, or that asked unnecessarily is a \emph{verified failure}. Verified failures are attributed by type: coverage and grounding failures to IGA, non-separating or unnecessary questions to CCA, and inconsistent answers to EFA. A process-review prompt handles residual cases ($91\%$ agreement with manual labels; Appendix~\ref{app:attribution}). Verified trajectories feed two loops, scheduled by whether the assistant is currently serving requests (Fig.~\ref{fig:overview}, top).

\textbf{Fast Loop: Intent-specific Skills.}
Verified failures form support set $\mathcal D^{\mathrm{sup}}_g$, from which an evolver updates $\mathcal S_{g+1}=\mathcal S_g\cup E_{\mathrm{skill}}(\mathcal D^{\mathrm{sup}}_g,\mathcal S_g)$. Each agent-specific skill records an ambiguity trigger, candidate topology (\eg, one object encountered twice), decisive contrast, corrected action, anti-pattern, and two regression tests against unnecessary asking. Up to three skills, retrieved by trigger--topology embedding similarity, immediately enter the responsible agent's prompt. For example, a CCA skill contrasts repeated encounters through the user's visible action rather than their order; Appendix~\ref{app:skill-example} provides the worked example.

\textbf{Slow Loop: Agent-attributed Process Rewards.}
During idle periods, a query buffer $\mathcal D^{\mathrm{qry}}$ of 128 verified trajectories tagged with their skill generation, including successes and failures, triggers adapter updates. Each trajectory $\xi$ receives
\begin{equation}
R_{\mathrm{proc}}(\xi)
=r_{\mathrm{final}}
+\alpha\,r_{\mathrm{ground}}
+\beta\sum_t\Delta_t^{\star}
-\lambda\sum_t C_x(c_t)
-\mu\,r_{\mathrm{invalid}},
\label{eq:intent-prm}
\end{equation}
where $r_{\mathrm{final}}$ combines correctness and verifier verdict, $r_{\mathrm{ground}}$ combines $U_{\mathrm{grd}}$ and target coverage, $C_x$ is realized respondent cost, and $r_{\mathrm{invalid}}$ counts rule violations. Target progress $\Delta_t^\star=\log p_{t+1}(z^\star)-\log p_t(z^\star)$ rewards belief moving toward the intended interpretation. A fixed candidate scope prevents reward from candidate removal alone (Appendix~\ref{app:reward-details}).

For each failed step, the attributed agent samples $G=8$ alternatives from its logged prompt, including its original skills. The evaluation simulator completes them into trajectories; a reward model predicts $R_{\mathrm{proc}}$ (held-out Spearman $0.83$). GRPO~\citep{shao2024deepseekmath} updates only that agent's LoRA adapter, with a KL penalty toward frozen Base. Promotion requires improvement on the target failure type without degrading overall accuracy, unnecessary asking, or mean cost on a fixed validation probe; otherwise the update is rolled back (Appendix~\ref{app:promotion}).

\section{Experiments}
\label{sec:experiments}

\providecommand{\up}[1]{\textsuperscript{\tiny\color{teal!75!black}$\uparrow$#1}}
\providecommand{\dn}[1]{\textsuperscript{\tiny\color{teal!75!black}$\downarrow$#1}}
\providecommand{\wo}[1]{\textsuperscript{\tiny\color{gray}$\downarrow$#1}}
\providecommand{\wu}[1]{\textsuperscript{\tiny\color{gray}$\uparrow$#1}}

\setlength{\textfloatsep}{8pt plus 2pt minus 2pt}
\setlength{\floatsep}{6pt plus 2pt minus 2pt}
\setlength{\intextsep}{6pt plus 2pt minus 2pt}
\setlength{\abovecaptionskip}{3pt}
\setlength{\belowcaptionskip}{2pt}

Our experiments answer four questions. \textbf{Q1:} Does targeted clarification improve intent-aligned and grounded answering over direct and interactive baselines (Sec.~\ref{sec:main-results})? \textbf{Q2:} Does grounding the alternatives across time help most when the competing referents are separated in time (same object seen twice / similar objects seen at different moments) rather than visible at once (Sec.~\ref{sec:temporal-diagnostic})? \textbf{Q3:} Are the gains obtained with limited user effort (Sec.~\ref{sec:efficiency})? \textbf{Q4:} Does learning from interaction reduce future errors beyond simpler adaptation (Secs.~\ref{sec:main-results} and~\ref{sec:learning-dynamics})?

\vspace{-1mm}
\subsection{Experimental Setup}
\label{sec:experimental-setup}
\vspace{-1mm}

\paragraph{Benchmarks.}\label{sec:evaluation-benchmarks} 
To facilitate our study, we augment two closest-aim egocentric QA datasets with intent-ambiguity labels, minimal supporting temporal intervals and spatial boxes.
BuddyVQA~\citep{qin2026buddyvqa}, our primary benchmark, with its causal streaming protocol, ego-deictic expressions, and history-dependent question chains; and EgoBlind-Intent~\citep{xiao2025egoblind}, with timestamped assistance questions from blind and visually impaired users, used for zero-shot transfer.
We also test transfer on two independent image-based clarification benchmarks, ClearVQA~\citep{jian2025clearvqa} and ContextClarify~\citep{cao2026clarifyanswer}, treating the image as a one-frame prefix with their official metrics and respondents.
More benchmark augmentation details are in Appendix~\ref{app:benchmarks}.

\textbf{Protocol and Metrics.} A model may access only frames observed before the question timestamp and the preceding dialogue; gold labels, resolved questions, intervals, and boxes are never inputs at prediction time; only the latter three and the judge verdict are revealed to learning methods, after a prediction is logged. Answer accuracy uses each benchmark's multi-reference LLM-as-a-judge protocol over all test inputs, with \textsc{Conditional} and \textsc{Abstain} counted as incorrect; temporal metrics are computed on grounding-eligible questions, spatial metrics and Acc@GQA~\citep{xiao2024nextgqa} (we further require correct answer with both T-IoU and S-IoU $\ge0.5$) on the box-annotated subset. Interactive evaluation additionally reports ambiguity-detection F1, ask precision/recall, clarification accuracy, unnecessary-ask and abstention rates, turns, question tokens, response time, latency, respondent cost, and the utility of Eq.~\ref{eq:referclaw-objective}. All numbers are means over three fixed stream orders, and all subset values of a row come from the same runs and checkpoint; the full protocol, the population of every metric, and confidence intervals are in Appendices~\ref{app:protocol} and~\ref{app:ci}.

\textbf{Baselines and Implementation.}\label{sec:baselines-implementation} We compare against direct MLLMs~\citep{qwen2026qwen35}, grounded and streaming VideoQA systems~\citep{meng2025openo3video,huang2026egvqa,di2025rekv,wang2025rga3,qin2026buddyvqa}, and interactive clarification methods~\citep{jian2025clearvqa,yang2026clarifier,cao2026clarifyanswer}. Framework-level baselines and all \modelname~variants use Qwen3.5-9B, under the same input protocol.
To avoid conflating clarification, extra information, and online learning, we evaluate in two regimes. 
\emph{Frozen inference} compares fixed systems, including \textbf{\modelname~(Base)} and its asking-disabled control, which retains grounding, rewriting, and verification. \emph{Learning from interaction} follows MetaClaw's online evaluation~\citep{xia2026metaclaw}: feedback follows each logged prediction and updates affect only later inputs. We compare \textbf{\modelname~(Skills)} (fast skill adaptation), \textbf{\modelname~(Full)} (adding agent-attributed GRPO, Eq.~\ref{eq:intent-prm}), \emph{Reflexion-style memory}~\citep{shinn2023reflexion}, and \emph{MetaClaw-style} adaptation (generic skills, all-adapter updates, final-answer reward). All share the initial system, feedback, stream, and update budget.
Details are in Appendix~\ref{app:experimental-details}.

\vspace{-1mm}
\subsection{Main Results}
\label{sec:main-results}
\vspace{-1mm}

\begin{table}[t]
\centering
\footnotesize
\setlength{\tabcolsep}{4pt}
\renewcommand{\arraystretch}{0.9}
\caption{Ego-streaming QA accuracy (\%) on Augmented BuddyVQA. $\dagger$: published results under the original protocol and split (not directly comparable, Appendix~\ref{app:datasets}). All other rows are comparable.
}
\label{tab:buddy-main}
\begin{adjustbox}{max width=\textwidth}
\begin{tabular}{lrrrrrrr}
\toprule
Method & Recall & Recog. & Advis. & Scrut. & Chain & Deictic & Overall \\
\midrule
Human$^\dagger$ & 92.4 & 92.1 & 91.9 & 92.0 & -- & -- & 92.1 \\
InternVL3.5-8B$^\dagger$~\citep{wang2025internvl35} & 54.1 & 51.3 & 44.7 & 46.1 & 47.2 & 43.1 & 48.8 \\
Flash-VStream$^\dagger$~\citep{zhang2025flashvstream} & 52.2 & 51.9 & 44.6 & 46.7 & -- & -- & 50.8 \\
ReKV (LLaVA-OV-7B)$^\dagger$~\citep{di2025rekv} & 54.6 & 54.1 & 44.9 & 44.4 & 50.9 & 48.6 & 52.7 \\
MyBuddy (Qwen3.5-9B)$^\dagger$~\citep{qin2026buddyvqa} & 59.9 & 60.8 & 55.8 & 54.2 & 57.3 & 58.9 & 60.7 \\
\midrule
\multicolumn{8}{l}{\emph{\color{lightgray} Frozen inference (Qwen3.5-9B)}} \\
Qwen3.5-9B~\citep{qwen2026qwen35} & 52.8 & 55.9 & 46.6 & 49.1 & 46.8 & 47.9 & 51.4 \\
MyBuddy (our reproduction)~\citep{qin2026buddyvqa} & 62.0 & 62.3 & 53.4 & 57.9 & 59.8 & 57.9 & 59.2 \\
ClearVQA~\citep{jian2025clearvqa} & 57.7 & 59.8 & 50.9 & 53.2 & 51.7 & 54.0 & 55.7 \\
Plug-and-Play Clarifier~\citep{yang2026clarifier} & 59.1 & 61.3 & 52.9 & 55.2 & 53.9 & 57.1 & 57.4 \\
Clarify-or-Answer~\citep{cao2026clarifyanswer} & 60.4 & 62.1 & 53.8 & 56.5 & 55.2 & 58.3 & 58.5 \\
\cmidrule(lr){1-8}
\modelname~(Base, asking disabled) & 62.0\up{1.6} & 62.8\up{0.7} & 56.3\up{2.5} & 60.2\up{3.7} & 58.6\up{3.4} & 58.2\wo{0.1} & 60.5\up{2.0} \\
\modelname~(Base) & \textbf{67.7\up{5.7}} & \textbf{68.6\up{5.8}} & \textbf{62.3\up{6.0}} & \textbf{65.4\up{5.2}} & \textbf{66.8\up{8.2}} & \textbf{69.0\up{10.8}} & \textbf{66.2\up{5.7}} \\
\midrule
\multicolumn{8}{l}{\emph{\color{lightgray} Learning from interaction (same stream, feedback, and update budget)}} \\
Base + Reflexion-style memory~\citep{shinn2023reflexion} & 68.9 & 69.7 & 63.5 & 66.7 & 68.0 & 70.1 & 67.4 \\
Base + MetaClaw-style adaptation~\citep{xia2026metaclaw} & 71.1 & 71.7 & 65.8 & 68.7 & 70.2 & 72.3 & 69.5 \\
\cmidrule(lr){1-8}
\modelname~(Skills) & 71.6\up{0.5} & 72.2\up{0.5} & 66.4\up{0.6} & 69.4\up{0.7} & 71.0\up{0.8} & 73.1\up{0.8} & 70.1\up{0.6} \\
\modelname~(Full) & \textbf{75.7\up{4.1}} & \textbf{76.4\up{4.2}} & \textbf{71.0\up{4.6}} & \textbf{73.3\up{3.9}} & \textbf{75.4\up{4.4}} & \textbf{77.5\up{4.4}} & \textbf{74.3\up{4.2}} \\
\bottomrule
\end{tabular}
\end{adjustbox}
\end{table}

\textbf{Frozen Inference (Q1):} Tab.~\ref{tab:buddy-main} compares answer accuracy under the streaming protocol. Frozen \modelname~reaches 66.2\%, versus 58.5\% for Clarify-or-Answer. Disabling asking while retaining grounding and verification lowers accuracy to 60.5\%, giving a 5.7-point matched clarification gain. Improvements also appear on deictic and history-dependent questions: Base reaches 69.0\% and 66.8\%, respectively, versus 58.3\% and 55.2\% for Clarify-or-Answer. The category breakdown and published reference results appear in Tab.~\ref{tab:buddy-main}.
\textbf{Learning From Interaction (Q4):} Starting from Base, intent-specific skills raise accuracy to 70.1\%, and Full reaches 74.3\%, versus 67.4\% for Reflexion-style memory and 69.5\% for MetaClaw-style adaptation. These comparisons measure adaptation under the same feedback and update budget. Sec.~\ref{sec:learning-dynamics} further tests whether improvements also hold on an unchanged probe set.

\begin{figure}[t]
\centering
\includegraphics[width=\linewidth,height=0.35\linewidth]{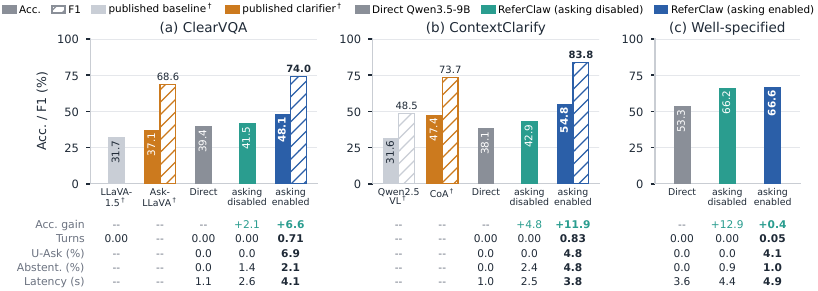}
\vspace{-5mm}
\caption{Clarification \vers No Clarification (\modelname~uses frozen Base; $\dagger$: published). Solid bars: accuracy; hatched bars: F1. Rows under each panel: accuracy gain over the preceding bar and interaction statistics.}
\label{fig:clarification-benchmarks}
\vspace{-2mm}
\end{figure}

\begin{wraptable}{r}{0.44\linewidth}
\vspace{-3.5mm}
\centering
\small
\setlength{\tabcolsep}{4pt}
\caption{Results on EgoBlind-Intent (\%). }
\label{tab:egoblind-main}
\resizebox{\linewidth}{!}{
\begin{tabular}{lrrrr}
\toprule
Method & T-mIoU & S-mIoU & Ans. & Acc@GQA \\
\midrule
Open-o3-Video & 44.7 & 41.2 & 54.0 & 27.0 \\
EG-Reasoner & 46.5 & 39.8 & 54.8 & 27.6 \\
\modelname~(Skills) & 48.9 & 41.6 & 58.8 & 30.1 \\
\modelname~(Full) & \textbf{50.1} & \textbf{42.6} & \textbf{60.7} & \textbf{31.6} \\
\bottomrule
\end{tabular}
}
\vspace{-3mm}
\end{wraptable}
\textbf{Generalization:} Fig.~\ref{fig:clarification-benchmarks} addresses two concerns: whether clarification helps beyond our own annotations, and whether it disrupts questions whose intent is already clear. F1 measures ambiguity detection on ClearVQA and the ask/answer decision on ContextClarify; Turns counts questions per input, U-Ask is the unnecessary-ask rate, and latency excludes respondent time. Without training skills or adapters on either image benchmark (protocols and populations in Appendix~\ref{app:clarification-benchmarks}), enabling clarification raises ClearVQA score from 41.5\% to 48.1\% and ContextClarify accuracy from 42.9\% to 54.8\%. These image-based tests assess clarification transfer, not temporal disambiguation. On well-specified BuddyVQA questions, accuracy changes by $+0.4$ points and the assistant asks on 4.1\% of inputs. Confidence intervals are in Appendix~\ref{app:ci}.
\textbf{Joint Grounding and Cross-domain Transfer:} Correct answers may not be supported by the intended visual evidence. We therefore also evaluate Acc@GQA on our augmented benchmarks (Tab.~\ref{tab:grounded}, Appendix~\ref{app:extended-grounding}). Frozen Base reaches 36.2\% on BuddyVQA versus 35.0\% for EG-Reasoner~\citep{huang2026egvqa}, while Full reaches 41.8\%. Under zero-shot transfer to EgoBlind-Intent (Tab.~\ref{tab:egoblind-main}, all thresholds, skills, and adapters frozen; full results for both benchmarks in Tab.~\ref{tab:grounded}), Full keeps a 4.0-point Acc@GQA and 5.9-point answer-accuracy advantage over EG-Reasoner, while both metrics drop by 10--12 points relative to BuddyVQA; the drop concentrates on blurred or off-center instances, whereas clarification quality transfers almost fully (Appendix~\ref{app:egoblind-breakdown}), so the gap lies in perception rather than in the clarification policy.

\vspace{-1mm}
\subsection{Ablation Study}
\label{sec:ablation-study}
\vspace{-1mm}

\begin{table}[t]
\centering
\footnotesize
\setlength{\tabcolsep}{3pt}
\renewcommand{\arraystretch}{0.9}
\caption{Agent and communication ablation on Augmented BuddyVQA; each row removes one agent output or cross-agent message with all else fixed. Clar.: clarification attempts after which the EFA selects the target intent; Ask: fraction of candidate-bearing inputs with a question; Cost: $\sum_t C_x(c_t)$ per asked input. The last four columns average the per-input terms of Eq.~\ref{eq:referclaw-objective} over candidate-bearing inputs, with Utility $=\overline{U}_{\mathrm{ans}}+0.5\,\overline{U}_{\mathrm{grd}}-\overline{\Sigma C}$ (Appendix~\ref{app:protocol}). Superscripts: change relative to the full model.}
\label{tab:agent-ablation}
\begin{adjustbox}{max width=\textwidth}
\begin{tabular}{lrrrrrrrrr}
\toprule
Variant & Ans. & Acc@GQA & Clar. & Ask & Cost & $\overline{U}_{\mathrm{ans}}$ & $\overline{U}_{\mathrm{grd}}$ & $\overline{\Sigma C}$ & Utility \\
\midrule
\modelname~(Full) & \textbf{74.3} & \textbf{41.8} & \textbf{81.4} & \textbf{38.7} & \textbf{0.42} & \textbf{0.732} & \textbf{0.328} & \textbf{0.163} & \textbf{0.733} \\
\midrule
\multicolumn{10}{l}{\emph{Intention Grounding Agent}} \\
$-$ IGA candidates (global video to CCA/EFA) & 68.0\wo{6.3} & 30.5\wo{11.3} & 68.2\wo{13.2} & 42.0\wu{3.3} & 0.48\wu{0.06} & 0.666 & 0.264 & 0.202 & 0.596\wo{0.136} \\
IGA temporal only ($-$ spatial tracks) & 72.1\wo{2.2} & 37.1\wo{4.7} & 75.6\wo{5.8} & 40.0\wu{1.3} & 0.44\wu{0.02} & 0.714 & 0.302 & 0.176 & 0.689\wo{0.044} \\
\midrule
\multicolumn{10}{l}{\emph{Evidence-Fused Answering Agent}} \\
EFA on global video ($-$ selected evidence) & 71.0\wo{3.3} & 35.2\wo{6.6} & 81.1\wo{0.3} & 38.7 & 0.42 & 0.701 & 0.289 & 0.163 & 0.683\wo{0.050} \\
$-$ resolved-intent rewrite and history in EFA & 71.4\wo{2.9} & 38.8\wo{3.0} & 73.5\wo{7.9} & 38.7 & 0.42 & 0.704 & 0.315 & 0.163 & 0.699\wo{0.034} \\
$-$ answer--intent--evidence verifier & 72.7\wo{1.6} & 39.7\wo{2.1} & 81.4 & 38.7 & 0.42 & 0.714 & 0.319 & 0.163 & 0.711\wo{0.022} \\
\midrule
\multicolumn{10}{l}{\emph{Cross-agent communication and control}} \\
CCA without IGA candidates & 70.1\wo{4.2} & 36.3\wo{5.5} & 65.2\wo{16.2} & 45.0\wu{6.3} & 0.51\wu{0.09} & 0.690 & 0.293 & 0.230 & 0.607\wo{0.126} \\
EFA without CCA-updated posterior & 70.8\wo{3.5} & 37.3\wo{4.5} & 69.8\wo{11.6} & 38.7 & 0.42 & 0.702 & 0.304 & 0.163 & 0.691\wo{0.041} \\
$-$ controller and posterior stopping & 69.9\wo{4.4} & 36.9\wo{4.9} & 71.2\wo{10.2} & 52.1\wu{13.4} & 0.66\wu{0.24} & 0.682 & 0.297 & 0.344 & 0.486\wo{0.247} \\
\bottomrule
\end{tabular}
\end{adjustbox}
\vspace{-2mm}
\end{table}

\paragraph{Agents and Shared Candidates:} Tab.~\ref{tab:agent-ablation} evaluates \modelname~(Full) under matched stream order, backbone, interaction limit, and training budget. Grounded candidates support both intent resolution and clarification: replacing IGA's candidate evidence with global video lowers Acc@GQA by 11.3 points, while withholding its candidates from CCA reduces clarification success by 16.2 points. The EFA ablations show that selected evidence, the resolved-intent rewrite and history, and verification each contribute to answer quality. Posterior-based stopping also improves accuracy while reducing respondent cost. Together, these results support sharing grounded alternatives across question selection and answering. Appendix~\ref{app:design-cases} (Fig.~\ref{fig:design-cases}) maps the four core designs to case types: grounded candidates and contrastive clarification account for the gains on referents separated in time and for staying silent when history resolves the reference, evidence-fused answering for questions that require reasoning after the referent is resolved, and learning from interaction for failure types that recur along the stream, so the designs are complementary rather than redundant.
\textbf{Isolating Clarification: }
Disabling asking while holding the entering model state and history fixed lowers Full accuracy from 74.3\% to 66.9\%, measuring the immediate benefit of clarification. A separate online run that never asks reaches 64.8\%, reflecting the combined effects on answering and subsequent learning. A top-two clarification heuristic reaches 71.0\% at higher respondent cost, supporting selective question choice. Removing the ambiguity gate or cost penalty increases asking frequency, while generic questions reduce clarification success from 81.4\% to 64.9\%. Full controls and further ablations are in Appendices~\ref{app:controls} and~\ref{app:fine-ablation}.

\vspace{-1mm}
\subsection{Further Analysis}
\label{sec:further-analysis}
\vspace{-1mm}

\begin{figure}[t]
\centering
\begin{minipage}[t]{0.49\linewidth}
    \centering
    \includegraphics[width=0.97\linewidth, height=0.5\linewidth]{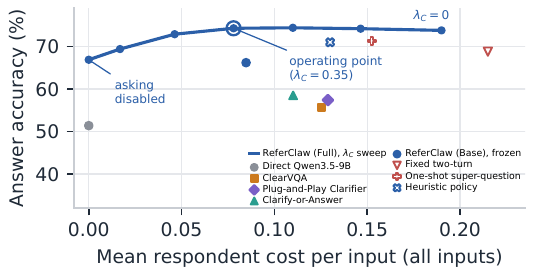}
    \vspace{-2mm}
    \caption{Accuracy--respondent-cost trade-off on Augmented BuddyVQA; the curve sweeps $\lambda_C$ of \modelname~(Full).}
    \label{fig:analysis-a}
\end{minipage}\hfill
\begin{minipage}[t]{0.49\linewidth}
    \centering
    \includegraphics[width=0.97\linewidth, height=0.5\linewidth]{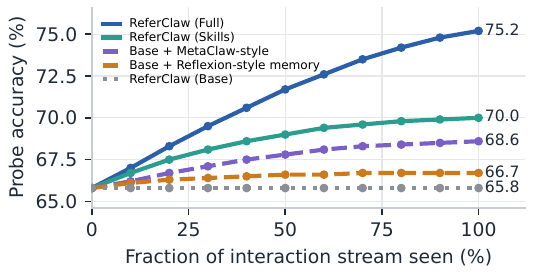}
    \vspace{-2mm}
    \caption{Fixed-probe accuracy after each 10\% of the stream under identical feedback and update budgets.}
    \label{fig:analysis-b}
\end{minipage}
\\[2mm]
\begin{minipage}[t]{0.49\linewidth}
    \centering
    \includegraphics[width=0.97\linewidth, height=0.5\linewidth]{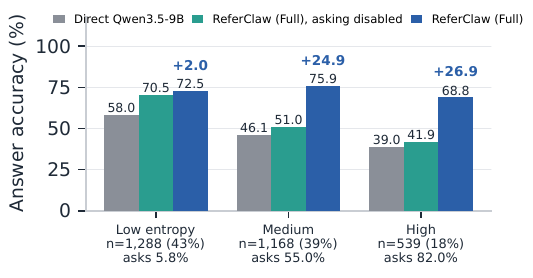}
    \vspace{-2mm}
    \caption{Accuracy by initial candidate entropy on candidate-bearing questions (labels: ask rate and share per bin).}
    \label{fig:analysis-c}
\end{minipage}\hfill
\begin{minipage}[t]{0.49\linewidth}
    \centering
    \includegraphics[width=0.97\linewidth, height=0.5\linewidth]{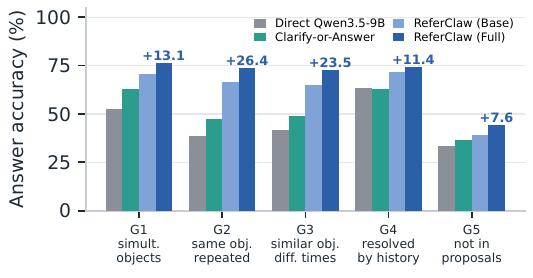}
    \vspace{-2mm}
    \caption{Accuracy by ambiguity group G1--G5 (Sec.~\ref{sec:temporal-diagnostic}); labels: \modelname~(Full) minus Clarify-or-Answer.}
    \label{fig:analysis-d}
\end{minipage}
\vspace{-3mm}
\end{figure}

\textbf{1. Direct Test of Temporal Ambiguity (Q2).}\label{sec:temporal-diagnostic} Fig.~\ref{fig:analysis-d} tests whether the benefit is concentrated in temporal ambiguity by partitioning the referentially underspecified questions into five groups (sizes and assignment rules in Appendix~\ref{app:annotation}). Frozen Base exceeds Clarify-or-Answer by 19.1 points for repeated encounters with the same object (G2) and 16.1 for similar objects at different times (G3), versus 7.8 for objects visible simultaneously (G1); the Full-minus-Clarify-or-Answer margins annotated in the figure follow the same pattern. When history already resolves the ambiguity (G4), the baselines still ask on about a quarter of the inputs (27.9\% for Plug-and-Play Clarifier, 24.4\% for Clarify-or-Answer), whereas Base asks on 7.0\% and Full on 6.5\%; all methods degrade when the target is absent from the proposals (G5). The result visualization (Fig.~\ref{fig:qual-example}, Appendix~\ref{app:traces}) shows both effects on one episode: \modelname~names two temporally separated encounters in one question and answers the history-resolved follow-up without asking.
A separate paired-intent test presents identical video and questions with two mutually exclusive target answers, changing only the respondent's clarification. Full answers both presentations correctly on 71.3\% of pairs (Base: 64.2\%), versus 38.8\% for Clarify-or-Answer and 33.3\% for Plug-and-Play Clarifier; deterministic direct answering scores zero because it receives identical inputs. This tests whether answers follow the clarified intent rather than a fixed interpretation; clarification-quality metrics on the ambiguous set and example traces are in Appendices~\ref{app:clarification-quality} and~\ref{app:traces}.

\textbf{2. Efficiency and Respondent Effort (Q3).}\label{sec:efficiency} Fig.~\ref{fig:analysis-a} averages respondent cost over all inputs; the Full curve sweeps $\lambda_C$ from $0$ (right) to asking disabled (left). Filled markers denote frozen systems and hollow markers Full-based policy controls. Base outperforms frozen clarification baselines at lower cost. Full asks on 18.6\% of inputs, averaging 0.22 questions per input. Its questions average 20.6 tokens and take humans 5.1\,s to answer, against 24.0 tokens and 6.9\,s for Clarify-or-Answer. Removing the cost penalty more than doubles cost without improving accuracy; a fixed two-turn policy also costs more and performs worse. Gains over the matched asking-disabled replay are larger for high-entropy candidates (Fig.~\ref{fig:analysis-c}; Appendix~\ref{app:model-analysis}). Timing and cost components are in Appendix~\ref{app:efficiency}. Human-respondent results are reported in Appendix~\ref{app:human}.

\textbf{3. Learning Dynamics (Q4).}\label{sec:learning-dynamics} On an unchanged held-out probe, Skills improves over Base by 4.2 points and Full by 9.4, compared with 0.9 for Reflexion-style memory and 2.8 for MetaClaw-style adaptation (Fig.~\ref{fig:analysis-b}). Full also cuts failure recurrence from 34.2\% to 10.7\% with positive backward transfer ($+$4.1), versus 21.3\% and $+$1.1 for MetaClaw-style adaptation. Targeted ablations further show that a generic skill schema reduces answer accuracy by 1.8 points and updating all adapters rather than the attributed agent reduces it by 2.0 (Appendix~\ref{app:fine-ablation}). 
Since the MetaClaw-style variant alters the skill schema, the adapter-update scope, and the reward jointly, its gap to Full confounds the three factors; the single-factor ablations above disentangle them and attribute a gain to each. Learning curves, recurrence rates, and backward-transfer results are reported in Appendix~\ref{app:learning-dynamics}, and a manual audit of residual errors is provided in Appendix~\ref{app:model-analysis}.

\section{Conclusion}
\label{sec:conclusion}
We studied intent ambiguity in egocentric assistance, where video and dialogue evidence supports multiple possible referents. \modelname~explicitly represents competing interpretations with spatio-temporal evidence, allowing the same grounded candidates to determine whether to clarify, what to ask, and which evidence to use. Three cooperating agents make these decisions under a fixed interaction budget, while verified feedback is distilled into intent-specific skills and agent-attributed process rewards to reduce recurring errors. On augmented ego-streaming assistance benchmarks, \modelname~asks more useful clarification questions than strong interactive baselines, improves intent-aligned and jointly grounded answers with lower user effort, and provides the largest gains when competing referents are temporally separated. The learned skills and parameters further improve fixed probes and reduce repeated failures, while the frozen system transfers to two independent image-based clarification benchmarks without degrading well-specified questions.

\section*{Ethics Statement}
This work adheres to the ICLR Code of Ethics. All video and image data come from publicly released benchmarks (BuddyVQA, EgoBlind, ClearVQA, and ContextClarify) and are used under their licenses. The user study (Appendix~\ref{app:human-calibration}) involved adult volunteers who gave informed consent, answered clarification questions about public clips, and were compensated; only anonymized responses and response times were recorded. Because \modelname~targets assistive use, including questions from blind and visually impaired users, an answer to the wrong interpretation could mislead a user who cannot verify the evidence, and clarification imposes effort. The system therefore grounds answers in verifiable evidence, abstains or gives conditional answers when the intent cannot be resolved, and bounds the number, length, and cost of questions; it remains a research prototype and should not be deployed for safety-critical assistance without validation.

We have taken care to avoid any possible biases or discriminatory outcomes in our research process. No personally identifiable information was used, and no experiments were conducted that could raise privacy or security concerns. We are committed to maintaining transparency and integrity throughout the research process.

\section*{Reproducibility Statement}
We have made every effort to ensure that the results presented in this paper are reproducible. All code, prompts, trained LoRA adapters, the skill library, and our augmented annotations (intent-ambiguity labels, target intents, supporting intervals, evidence-frame boxes, diagnostic groups, and the paired-intent set) will be provided upon acceptance. The method is specified in Sec.~\ref{sec:method} and Appendix~\ref{app:method-details} (controller transition table, response likelihood and cost, process reward, promotion rule); all hyperparameters are listed in Tab.~\ref{tab:referclaw-hparams} and the hardware and optimization setup in Appendix~\ref{app:experimental-details}. The evaluation protocol, including what the assistant and the simulated user receive, when feedback is revealed, and the population of every metric, is documented in Appendix~\ref{app:protocol}, and all prompts are given verbatim in Appendix~\ref{app:prompts-templates}. All benchmarks and backbone or tool models are publicly available. Human-respondent results depend on the recruited participants and are reported separately.
We believe these measures will enable other researchers to reproduce our work and further advance the field.

\bibliography{iclr2027_conference}
\bibliographystyle{iclr2027_conference}

\appendix
\providecommand{\FloatBarrier}{}
\section{LLM Usage}
LLMs were used to aid in the writing and polishing of the manuscript. Specifically, we used an LLM to refine the language, improve readability, and ensure clarity across various sections of the paper. The model helped with tasks such as rephrasing sentences, checking grammar, and improving the overall flow of the text.

It is important to note that the LLM was not involved in the ideation, research methodology, or experimental design. All research concepts, ideas, and analyses were developed and conducted by the authors. The contributions of the LLM were solely focused on improving the linguistic quality of the paper, with no involvement in the scientific content or data analysis.

The authors take full responsibility for the content of the manuscript, including any text generated or polished by the LLM. We have ensured that the LLM-generated text adheres to ethical guidelines and does not contribute to plagiarism or scientific misconduct.

\FloatBarrier
\section{Method Details}
\label{app:method-details}

\FloatBarrier
\subsection{Module Instantiation and Candidate Grounding}
\label{app:candidates}

IGA, CCA, and EFA share a Qwen3.5-9B backbone but use separate LoRA adapters, which keeps the visual representation fixed while allowing agent-specific updates. IGA uses the released GroundVQA checkpoint for coarse temporal proposals, Grounding DINO 1.5 Pro~\citep{ren2024groundingdino15} for open-vocabulary boxes, and SAM~2 Hiera-Large~\citep{ravi2024sam2} to propagate the selected box through the retrieved interval. CCA and EFA operate on the resulting evidence cards and Qwen3.5-9B video tokens. BGE-M3~\citep{chen2024bgem3} encodes skills and trajectories for cosine retrieval. The process reward model initializes from Qwen3.5-9B, adds a scalar value head, and is trained to regress the computed reward of Eq.~\ref{eq:intent-prm} on query trajectories with feedback (held-out Spearman $0.83$, MAE $0.09$ on a $[0,1]$-scaled reward). The \emph{predictive respondent model} used by CCA to estimate response likelihoods is the Qwen3.5-9B backbone under the predictive-respondent prompt of Appendix~\ref{app:prompts-templates}; it never receives the annotated target intent. The \emph{evaluation simulator} that generates the actual responses during testing and learning rollouts is Qwen3-32B~\citep{yang2025qwen3}; it receives the hidden target intent and its evidence but never the system posterior, candidate scores, or reference answer. Eight samples per $(c,z_i)$ pair estimate each response-class distribution.

\FloatBarrier
\subsection{Response Likelihood and Respondent Cost}
\label{app:cost-details}

\paragraph{Response Classes and Likelihood.} Free-text responses are assigned to the nearest response class (a partition cell, ``other/none'', \emph{uncertain}, or invalid/missing) by BGE-M3 embedding similarity. For the LLM respondent, the likelihood $p_{\mathrm L}(y\mid c,z_i,h_t)$ is the class frequency over eight samples of the predictive respondent model per $(c,z_i)$ pair, with additive smoothing $\delta=0.05$ so that no class receives zero probability. For the human respondent, $p_{\mathrm H}=(1-\kappa_{\mathrm H}-f_{\mathrm H})\,p_{\mathrm L}+\kappa_{\mathrm H}\,\mathrm{Unif}+f_{\mathrm H}\,q_{\mathrm F}$, where $q_{\mathrm F}$ is a normalized distribution over the failure classes, with $\kappa_{\mathrm H}$, $f_{\mathrm H}$, and $q_{\mathrm F}$ fitted as described in Appendix~\ref{app:human-calibration}. For both respondent types the failure-class likelihood depends on the question only, not on $z_i$, so a failure response leaves the belief unchanged under Eq.~\ref{eq:posterior-update}, which is also what the controller executes. The uncovered pseudo-candidate $z_\varnothing$ has the likelihood $p_x(y\mid c,z_\varnothing,h_t)=(1-f(c))\,\mathbb I[y=\text{other/none}]+f(c)\,q_{\mathrm F}(y)$, where $f(c)$ is the candidate-independent failure mass of the question: a user whose intent is not among the displayed candidates selects ``other/none'' unless the response fails. Entropy, expected information gain, and the posterior update are all computed over the $n_t+1$ elements $\{z_1,\dots,z_{n_t},z_\varnothing\}$.

\paragraph{Cost Components.} For a human respondent, $\overline E_{\mathrm H}$ averages the normalized question length, the expected response length under the predictive respondent, and the choice complexity $\log_2(m(c)+1)/\log_2(M_{\max}+1)$; $\overline T_{\mathrm H}$ is the response time predicted by a linear regression on question length and option count fitted on the calibration half of the user study ($\overline T_{\mathrm H}=1.0+0.12\,|c|+0.55\,m(c)$ seconds); and $\overline F_{\mathrm H}$ is the predicted mass of the invalid, missing, and uncertain response classes. For an LLM respondent, $\overline E_{\mathrm L}$ counts prompt and completion tokens, $\overline T_{\mathrm L}$ is measured serving latency, and $\overline F_{\mathrm L}$ is the predicted invalid-format or self-inconsistent mass. A failed response already yields little information gain; $\overline F_x$ additionally charges the wasted turn and the user's time. All components are normalized to $[0,1]$ with the 5th and 95th percentiles of the validation split, and the weights $(w_E,w_T,w_F)$ are fitted by grid search on validation utility separately for the two respondent types, with a shared $\lambda_C$ (Tab.~\ref{tab:referclaw-hparams}).

\FloatBarrier
\subsection{Human Response Model Calibration}
\label{app:human-calibration}

The 600-interaction user study (10 participants, 300 stratified ambiguous examples, each example answered under \modelname~and under Clarify-or-Answer by different participants) serves two purposes that we keep separate by example. The 300 examples are split once into a \emph{calibration} half (150 examples) and a \emph{held-out evaluation} half (150 examples). On the calibration half we use the \modelname~interactions to fit the human response model
\[
p_{\mathrm H}(y\mid c,z_i,h_t)=(1-\kappa_{\mathrm H}-f_{\mathrm H})\,\hat p_{\mathrm L}(y\mid c,z_i,h_t)+\kappa_{\mathrm H}\,\mathrm{Unif}(y)+f_{\mathrm H}\,q_{\mathrm F}(y),
\]
where $\hat p_{\mathrm L}$ is the predictive respondent's class distribution, $\mathrm{Unif}$ is uniform over all response classes, and $q_{\mathrm F}$ is a distribution supported on the three failure classes (\emph{uncertain}, \emph{invalid}, \emph{no response}) with proportions $(0.55,0.25,0.20)$ estimated from the study; each component is a normalized distribution, so $p_{\mathrm H}$ sums to one. ``Other/none'' is an ordinary answer class, not a failure class, because it is the correct response when the intended referent is missing from the options. The fitted weights are $\kappa_{\mathrm H}=0.18$ and $f_{\mathrm H}=0.06$. Response time is fitted on the same half as $\overline T_{\mathrm H}=1.0+0.12\,|c|+0.55\,m(c)$ seconds ($R^2=0.61$), with $|c|$ the question length in tokens and $m(c)$ the number of options. On the held-out half, the calibrated model lowers the Brier score of response prediction from $0.31$ (uncalibrated simulator frequencies) to $0.22$, and its predicted response times correlate with measured times at $r=0.74$. These parameters define the likelihoods and costs used when the respondent is human and the human-calibrated profile of the evaluation simulator; simulator-based likelihoods and costs use the simulator's own statistics. The human comparison in Appendix~\ref{app:human} uses only the held-out half for both methods; the Clarify-or-Answer interactions of the calibration half enter only the descriptive response-time statistics of Tab.~\ref{tab:efficiency-analysis}.

\FloatBarrier
\subsection{Interaction Control Algorithm}
\label{app:controller-alg}

The interaction controller of Sec.~\ref{sec:agents} is a deterministic state machine over the state $(t,\mathcal Z_t,p_t,h_t,\mathcal B_t,n_{\mathrm{rg}})$, where $\mathcal B_t$ is the set of banned partitions and $n_{\mathrm{rg}}\in\{0,1\}$ counts re-grounding passes. Tab.~\ref{tab:transitions} lists its transitions; at every turn the first row whose condition holds is executed, and the implementation, the pseudocode, and the main text share this table. All limits ($T_{\max}$, $L_{\max}$, $M_{\max}$, no repeated partitions) are enforced before generation, so every trajectory terminates within $T_{\max}$ turns. Two procedures are shared by all rows. \textsc{AnswerPath}$(i^\star)$ runs the EFA on candidate $i^\star$ and then the verifier: \emph{pass} returns the answer; \emph{revise} regenerates once and returns the answer if the second verdict is \emph{pass}, otherwise \textsc{Abstain}; \emph{abstain} returns \textsc{Abstain}. \textsc{Reground}$(\cdot)$ runs the IGA once more, with relaxed detector thresholds (initial state) or restricted to the constraint stated in the response (after ``other/none''); it may be invoked at most once per question ($n_{\mathrm{rg}}\le1$), new candidates absorb the mass of $z_\varnothing$ in proportion to their grounding scores, and an empty result triggers row T1 on the next evaluation.

\begin{table}[!htbp]
\centering
\footnotesize
\setlength{\tabcolsep}{4pt}\renewcommand{\arraystretch}{1.15}
\caption{Priority-ordered controller transitions for one user question. Rows are tested top to bottom at every turn $t$; the first satisfied row is executed. $i^{\star}=\arg\max_i p_t(z_i)$; $(1),(2)$ index the two most likely candidates.}
\label{tab:transitions}
\begin{tabular}{cp{0.25\linewidth}p{0.40\linewidth}p{0.20\linewidth}}
\toprule
& Condition & Action & Output / next state \\
\midrule
T1 & $\mathcal Z_t=\emptyset$ & if $n_{\mathrm{rg}}=0$: \textsc{Reground}, $n_{\mathrm{rg}}\gets1$, re-evaluate; else terminate & \textsc{Abstain} \\
T2 & $g_t=0$ \textbf{or} $\max_i p_t(z_i)\ge\tau_p$ & \textsc{AnswerPath}$(i^\star)$; if $g_t=0$ because all plausible candidates share an answer, EFA receives their pooled evidence and $\gamma$ is their summed mass & \textsc{Answer} or \textsc{Abstain} \\
T3 & $g_t=1$, $t<T_{\max}$, CCA returns $c_t^\star$ with $\operatorname{IG}_t-\lambda_C C_x>0$ and partition $\notin\mathcal B_t$ & ask $c_t^\star$; receive $y_t$; $h_{t+1}\gets h_t\cup\{(c_t^\star,y_t)\}$; then (a) $y_t$ uncertain, invalid, or missing: $p_{t+1}\gets p_t$, $\mathcal B_{t+1}\gets\mathcal B_t\cup\{\text{partition}(c_t^\star)\}$; (b) $y_t$ = other/none or unmatched constraint: \textsc{Reground}$(y_t)$ if $n_{\mathrm{rg}}=0$, else \textsc{Abstain}; (c) otherwise $p_{t+1}$ by Eq.~\ref{eq:posterior-update}, drop candidates with $p_{t+1}<0.02$; $t\gets t+1$ & continue at T1 \\
T4 & otherwise ($t=T_{\max}$, or no admissible question with positive utility) & if $\bar a_{(1)}\ne\bar a_{(2)}$ and $p_t(z_{(1)})-p_t(z_{(2)})<\tau_g$: \textsc{AnswerPath}$((1))$ and \textsc{AnswerPath}$((2))$, each labeled by its interpretation; else \textsc{AnswerPath}$(i^\star)$ & \textsc{Conditional} (if both branches pass; \textsc{Abstain} if either fails) or \textsc{Answer}/\textsc{Abstain} \\
\bottomrule
\end{tabular}
\end{table}

Three consequences of the table are used in the analysis. First, every formal answer, including each branch of a \textsc{Conditional} output, is produced by the same EFA-then-verifier path, so verifier statistics are comparable across output types. Second, a question is asked only through T3, so the number of questions per input is $N_{\mathrm{ask}}\le T_{\max}$ and a question can be asked only if $n_0\ge2$ (with $n_0=1$ the gate is closed). Third, uncertain and invalid responses never change the belief, which matches the candidate-independent failure-class likelihood used when scoring questions (Appendix~\ref{app:cost-details}), so the expected information gain computed before asking and the update executed after asking are consistent.

\FloatBarrier
\subsection{Failure Attribution Rules and Their Validation}
\label{app:attribution}

A verified failure is attributed to one agent by deterministic rules on the logged trajectory wherever possible: $z^\star$ absent from $\mathcal Z$, or present but grounded with $\mathrm{IoU}<0.5$ against $e^\star$, attributes to IGA (coverage or grounding failure); $z^\star$ present but placed in the same partition cell as the top competitor of the asked question, or a question asked when all plausible provisional answers already agreed, attributes to CCA (non-separating or unnecessary question); $z^\star$ selected with correct evidence but a wrong or unsupported answer attributes to EFA (inconsistent answer). Only trajectories that match none of these rules are attributed by the process-review prompt of Appendix~\ref{app:prompts-templates}. Two authors independently attributed 200 randomly sampled verified failures using the trajectory and feedback (inter-annotator agreement $\kappa=0.86$; disagreements adjudicated). The deterministic rules cover 84\% of the failures and agree with the adjudicated label in 94\% of covered cases; the process-review prompt handles the remaining 16\% with 78\% agreement, for a combined agreement of 91\%. Most residual disagreements concern trajectories in which the target intent was covered with borderline IoU ($0.4$--$0.5$) and the question also failed to separate it, \ie, cases with two plausible responsible agents.

\FloatBarrier
\subsection{Skill Schema and a Worked Example}
\label{app:skill-example}

Each skill stores an ambiguity trigger, a candidate topology, the decisive visual contrast, the corrected action, an anti-pattern, a retrieval key, and two regression tests (Appendix~\ref{app:prompts-templates} gives the evolver prompt and schema). Retrieval uses BGE-M3 cosine similarity between the current trigger--topology description and stored keys with threshold $0.45$; at most three skills are injected into the responsible agent's prompt. The following trace illustrates what is learned beyond the initial prompt, which already instructs CCA to contrast occurrences rather than repeat a noun. \emph{Failure:} for a mug picked up twice in the same kitchen within 40 seconds, CCA asked ``the first or the second time?''; the user answered \emph{uncertain} and the final answer was wrong. \emph{Feedback:} $z^\star$ was the second encounter, during which the user was pouring. \emph{Stored skill (CCA):} trigger ``same object, repeated encounters, same location, gap $<60$\,s''; corrected action ``contrast by the user's action visible in each track (holding \vers pouring), not by order or location''; anti-pattern ``ordinal or location contrast for repeated encounters at one place''. \emph{Later behavior:} on a subsequent kettle question with the same topology, CCA asked ``when you were filling it or when you were pouring?'', the response resolved the intent, and the answer was correct. Such action-contrast skills account for $23\%$ of the CCA skills learned on Augmented BuddyVQA.

\FloatBarrier
\subsection{Process Reward, Adapter Update, and Promotion}
\label{app:promotion}

\paragraph{Process Reward.}\label{app:reward-details}
The target-progress term $\Delta_t^{\star}=\log p_{t+1}(z^\star)-\log p_t(z^\star)$ of Eq.~\ref{eq:intent-prm} is evaluated on the candidate scope that existed \emph{before} turn $t$: candidates dropped or added after the response are re-inserted with their pre-update mass for the purpose of this term, so a change of scope cannot earn reward by itself, and $\Delta_t^{\star}$ is $0$ when $z^\star$ was not covered at turn $t$. Directly computed terms are $r_{\mathrm{ground}}$, $\sum_t C_x(c_t)$, $\sum_t\Delta_t^{\star}$, and $r_{\mathrm{invalid}}$; $r_{\mathrm{final}}$ combines the judge verdict with the verifier's verdict. The process reward model predicts $R_{\mathrm{proc}}$ for sampled alternative trajectories whose feedback is not yet available.

\paragraph{Adapter Update.}
When the query buffer holds 128 trajectories and no online request is being served, each failed step is re-executed: the attributed agent samples $G=8$ alternative structured outputs from its current adapter given the exact logged input, the process reward model scores each completed trajectory, and the adapter is updated with GRPO, \ie, group-normalized advantages under a clipped policy objective (clip ratio $0.2$) with a KL penalty (coefficient $0.02$) toward the frozen Base adapter. Only the attributed adapter changes. A candidate adapter is promoted only if, on a fixed validation probe of 500 ambiguous and 500 well-specified examples, it improves accuracy on its target failure type by at least one point, lowers overall probe accuracy by at most $0.5$ points, and does not raise the unnecessary-ask rate or mean cost by more than $0.01$; otherwise it is rolled back and the buffer is retained for the next consolidation window.

\FloatBarrier
\section{Benchmarks, Annotation Protocol, and Evaluation Details}
\label{app:benchmarks}

\FloatBarrier
\subsection{Augmented Benchmarks}
\label{app:datasets}

\paragraph{Augmented BuddyVQA.} BuddyVQA~\citep{qin2026buddyvqa} is our primary benchmark because its causal streaming protocol, ego-deictic expressions, and history-dependent question chains jointly exercise visual memory, referent tracking, and dialogue dependence. The original release contains 1,012 videos and 21,574 questions with a 704/308 video split for training and test; we annotate every question with an intent-ambiguity label, a minimal supporting temporal interval, and a box on the clearest evidence frame when the referent is spatially localizable (otherwise the instance is marked non-localizable). The original test split has 6,571 questions on the 308 test videos. We exclude 331 of them (5.0\%): 214 whose timestamp lies within the first 30 seconds of the video, so that no meaningful causal prefix exists, and 117 for which the two annotators could not agree on a reference answer under the annotated interpretation after adjudication. The resulting test split contains 6,240 questions (Recall 1,790, Recognition 1,650, Advisory 1,420, Scrutinization 1,380). The published BuddyVQA results quoted in Tab.~\ref{tab:buddy-main} ($\dagger$) were obtained by their authors under the original protocol on all 6,571 test questions; their type scores therefore do not satisfy the 6,240-question weighting, and we do not compare against them at the type level or compute differences to them. Of these, 1,398 are \emph{referentially underspecified} in wording; 1,197 of them (19.2\%; Recall 344, Recognition 318, Advisory 272, Scrutinization 263) remain intent-ambiguous given the prefix and the dialogue history and form the ambiguous population of every metric, whereas the other 201 are resolved by the history (group G4 below) and are labeled well-specified, giving 5,043 verified-unambiguous questions. 4,655 questions (74.6\%) are grounding-eligible, \ie, have an annotated supporting interval, and 3,914 of these additionally have a spatial box (741 are temporal-only).

\paragraph{Augmented EgoBlind-Intent.} EgoBlind~\citep{xiao2025egoblind} complements BuddyVQA with timestamped assistance questions from blind and visually impaired users under imperfect viewpoints, motion blur, and partial observation. We select 2,036 QA instances from the official test set and add the same ambiguity, interval, and box annotations. 612 instances are \emph{intent-relevant} because their ego-deictic, elliptical, or context-dependent wording requires explicit referent resolution; among these, 418 satisfy the stricter answer-changing criterion used for BuddyVQA and form the ambiguous subset on which clarification metrics are computed, and 1,447 instances are grounding-eligible. Because the two datasets differ in how much of the intent-relevant population is answer-changing, we never pool them. The benchmark is reserved for zero-shot transfer: no instance is used for prompt selection, threshold tuning, skill synthesis, or policy optimization.

\FloatBarrier
\subsection{Annotation Protocol}
\label{app:annotation}

\paragraph{Guidelines.} Annotators watched only the video prefix up to the question timestamp together with the preceding dialogue. For each question they (i) listed every entity, moment, or place that the question could refer to and that was visible in the prefix; (ii) for each interpretation wrote a resolved question and a short provisional answer from the video; (iii) marked the question intent-ambiguous if at least two interpretations were visually supported given the prefix \emph{and} the dialogue history and their provisional answers differed, and marked it history-resolved if the wording alone admitted several interpretations but a prior dialogue turn selected one of them; and (iv) annotated the minimal supporting interval and, when applicable, a single box on the clearest evidence frame for each interpretation (no box tracks are annotated; spatial IoU is therefore evaluated on that frame, Appendix~\ref{app:protocol}). The reference answer of the original dataset was shown only afterwards to select the annotated target intent among the enumerated interpretations; 3.1\% of BuddyVQA and 4.6\% of EgoBlind candidates for which the reference answer matched none of the enumerated interpretations were excluded from the ambiguous set and kept as ordinary questions. Two annotators labeled every instance independently (Cohen's $\kappa=0.78$ on BuddyVQA and $0.74$ on EgoBlind for the ambiguity label); disagreements were adjudicated by a third annotator, and 86\% of adjudicated resolved questions were judged equivalent to at least one annotator's proposal. Because intent is assigned retrospectively, all results should be read as accuracy with respect to the annotated target intent rather than the original user's intention.

\paragraph{Diagnostic Groups.} The groups of Sec.~\ref{sec:temporal-diagnostic} partition the 1,398 referentially underspecified questions; they are assigned in the following priority order so that every question belongs to exactly one group. G4 (201) if the dialogue history contains a prior mention that uniquely selects one interpretation; these questions are labeled well-specified and are used to measure unnecessary asking. Among the remaining 1,197 intent-ambiguous questions, G5 (131) if the annotated target interval has IoU $<0.3$ with all eight GroundVQA proposals, a tool-level condition shared by all methods; otherwise G1 (412) if the competing referents are distinct entities whose supporting intervals overlap, G2 (356) if the competing interpretations share the same entity track at disjoint intervals, and G3 (298) if the competing referents are distinct entities of the same category at disjoint intervals. The paired-intent set contains 240 videos for which two enumerated interpretations both have a valid reference answer in the original dataset (from different questions on the same video) and the two reference answers are judged mutually exclusive by the benchmark judge; we present the same ambiguous question twice with the respondent instructed with each target in turn, with the model, history, and random seed reset between the two presentations.

\FloatBarrier
\subsection{ClearVQA and ContextClarify}
\label{app:clarification-benchmarks}

ClearVQA~\citep{jian2025clearvqa} contains 4,025 test questions over single images in three ambiguity categories (referential 1,938, intent under-specification 1,095, spelling 992), an unambiguous contrast set for ambiguity detection, and a VQA-score protocol with and without clarification; we report VQA score, ambiguity-detection F1 on its contrast set, and its published LLaVA-1.5-13B and Ask-LLaVA-13B (SFT+DPO) numbers. ContextClarify~\citep{cao2026clarifyanswer} contains 275 ambiguous image--question pairs with human-verified clarifications and, for each, two contrast variants: a \emph{context-completed} variant in which the appended context resolves the ambiguity (gold decision \textsc{Answer}) and an \emph{irrelevant-context} variant in which plausible but non-resolving context is appended, so that the item remains ambiguous (gold decision \textsc{Clarify}). Its official test split has 42 ambiguous items and 84 variants (42 of each kind). We report end-to-end accuracy on the 42 ambiguous items, the \textsc{Clarify}-\vers-\textsc{Answer} F1 of the decision over all 126 test items (84 \textsc{Clarify} positives: the originals and the irrelevant-context variants), the unnecessary-ask rate on the 42 context-completed variants, and the published prompting and GRPO-CR numbers. For both benchmarks we treat the image as a one-frame prefix, disable temporal proposals, keep Grounding DINO for spatial candidates, and use each benchmark's official metrics and respondent protocol (ClearVQA's clarification-response simulator and ContextClarify's oracle context provider). No skill or adapter is trained on either benchmark.

\FloatBarrier
\subsection{Evaluation Protocol and Metric Definitions}
\label{app:protocol}

This subsection documents the protocol behind every number in the paper: what the assistant receives, what the simulated user knows, what is revealed after a prediction and to whom, when the assistant asks, answers, gives alternatives, or abstains, and which examples enter each metric.

\begin{table}[t]
\centering
\scriptsize
\setlength{\tabcolsep}{3pt}\renewcommand{\arraystretch}{0.95}
\caption{Population and success criterion of every reported metric on Augmented BuddyVQA (EgoBlind-Intent in parentheses). Cand.: candidate-bearing inputs; asked: inputs with at least one question.}
\label{tab:metric-populations}
\begin{tabular}{@{}p{0.235\linewidth}p{0.315\linewidth}p{0.43\linewidth}@{}}
\toprule
Metric & Averaged over & Success / definition \\
\midrule
\raggedright Ans.\ Acc.\ (Overall; type; Chain, Deictic) & \raggedright all 6,240 (2,036) inputs; that type or subset & \raggedright judge verdict w.r.t.\ target intent; \textsc{Conditional}/\textsc{Abstain} wrong \tabularnewline
\raggedright Ambiguous / well-specified acc. & \raggedright 1,197 / 5,043 (418 / --) & \raggedright Overall $=(1{,}197\,\mathrm{Amb}+5{,}043\,\mathrm{Well})/6{,}240$ \tabularnewline
\raggedright T-mIoU, T-R@.5, Ans.\ (Tab.~\ref{tab:grounded}) & \raggedright 4,655 (1,447) grounding-eligible & \raggedright mean T-IoU; T-IoU $\ge0.5$; judge verdict \tabularnewline
\raggedright S-mIoU, S-R@.5, Acc@GQA & \raggedright 3,914 (1,203) box-annotated & \raggedright S-IoU on annotated frame; S-IoU $\ge0.5$; correct and T-, S-IoU $\ge0.5$ \tabularnewline
\raggedright Amb-F1; Ask-P / Ask-R & \raggedright all inputs (1,197 ambiguous positives) & \raggedright F1 of the ambiguity flag; precision / recall of asking \tabularnewline
\raggedright Unnecessary-ask rate & \raggedright 5,043 well-specified (Fig.~\ref{fig:clarification-benchmarks}); 201 G4 (Sec.~\ref{sec:temporal-diagnostic}) & \raggedright at least one question asked \tabularnewline
\raggedright Clar.-Acc.\ (= Clar.\ Succ.); Intent-Acc. & \raggedright asked ambiguous, per method or variant (612 asked by all five); all 1,197 & \raggedright posterior mode, selected by EFA, is $z^\star$; answer addresses $z^\star$ \tabularnewline
\raggedright Ask rate; Turns (all / cand. / asked) & \raggedright all / cand.\ / asked inputs & \raggedright asked at least once; $N_{\mathrm{ask}}$ ($0$ if none, $\le T_{\max}$) \tabularnewline
\raggedright Q tokens; human time; GPU time, latency & \raggedright questions; study responses (else regression); all inputs & \raggedright tokens / question; s / response; s / input, respondent excluded \tabularnewline
\raggedright Cost (asked / all); Utility & \raggedright asked / all inputs; cand.\ inputs & \raggedright $\sum_t C_x(c_t)$ (``--'' if never asked); $U_{\mathrm{ans}}+0.5\,U_{\mathrm{grd}}-\sum_t C_x(c_t)$ \tabularnewline
\raggedright Abstention; paired-intent; probe & \raggedright all inputs; 240 pairs; 1,000 held-out & \raggedright \textsc{Conditional}/\textsc{Abstain}; both correct; verdict at checkpoint \tabularnewline
\bottomrule
\end{tabular}
\vspace{-2mm}
\end{table}

\paragraph{What Does the Assistant Receive?} Before producing any output for a test question, a method receives exactly four things: the causal video prefix $V_{\le s_q}$ decoded at 1\,fps up to the question timestamp, the question $q$, the preceding dialogue of the same episode (earlier questions, the method's own earlier answers, and earlier clarification exchanges), and the interaction budget ($T_{\max}=2$, $L_{\max}=24$, $M_{\max}=4$). Future frames, the gold ambiguity label, the annotated target intent $z^\star$, its resolved question, the supporting interval and box $e^\star$, the diagnostic-group label, and the reference answer are never inputs at prediction time, for any method and in any regime. On ClearVQA and ContextClarify the prefix is the single image.

\paragraph{What Does the Simulated User Know?} Two respondent roles are kept strictly separate (prompts in Appendix~\ref{app:prompts-templates}). The \emph{evaluation simulator} (Qwen3-32B; Llama-3.3-70B in Appendix~\ref{app:human}) answers the clarification questions that are actually asked during testing and learning rollouts. It receives the original question, the clarification history, the selected question with its options, the hidden target intent $z^\star$ as a resolved question together with a textual description of its evidence (object, place, and time of the intended encounter), and a respondent profile calibrated on the user study. It never receives the reference answer, the system posterior, candidate scores, the system's preferred option, or the video itself, and its rules forbid revealing more than the clarification asks. The \emph{predictive respondent} used internally by CCA to estimate response likelihoods is the assistant's own Qwen3.5-9B backbone prompted once per candidate $z_i$ ``as a user whose intent is $z_i$''; it never sees $z^\star$. Human participants in the user study see only the video prefix, the original question, the target intent description they are asked to hold, and the clarification question. On ClearVQA and ContextClarify we use the benchmarks' official respondents (ClearVQA's clarification-response simulator and ContextClarify's oracle context provider).

\paragraph{What Is Revealed after a Prediction, and Who Learns from It?} After a method has logged its final output for interaction $k$, the evaluator computes the judge verdict and grounding IoUs. Under the prequential protocol, the following \emph{verified feedback} is then made available: $z^\star$ (resolved question and evidence $e^\star$) and the binary judge verdict $U_{\mathrm{ans}}$. The reference answer string itself is not revealed, so no method can memorize answers. Only adapting conditions consume the feedback: \modelname~(Skills) and \modelname~(Full) with their learning-loop ablations and the online no-ask run, the Reflexion-style memory, and the MetaClaw-style variant. Feedback for interaction $k$ may affect interactions $k{+}1$ onward only; frozen methods (all baselines in the frozen block, \modelname~(Base), and both asking-disabled controls) receive the feedback stream but ignore it.

\paragraph{When Does the Assistant Ask, Answer, Give Alternatives, or Abstain?} The decision follows the controller transitions of Tab.~\ref{tab:transitions} (Appendix~\ref{app:controller-alg}): a method answers directly when the gate is closed or one candidate dominates (T2), asks only through T3, returns a \textsc{Conditional} output (one separately verified answer per top-two interpretation) when no further question is admissible but the two interpretations still disagree (T4), and \textsc{Abstain}s when no supported candidate or verified answer remains. \textsc{Conditional} and \textsc{Abstain} outputs are scored as incorrect in every accuracy column and reported separately as the abstention rate (Appendix~\ref{app:coverage}).

\paragraph{Answer Correctness and Grounding.} Answer accuracy uses each benchmark's multi-reference LLM-as-a-judge protocol; a stratified sample of 500 judgments was audited manually (97.2\% agreement). On ambiguous questions the judge compares the answer with the reference answer of the \emph{annotated target intent}, so an answer that is correct for a different interpretation counts as wrong. Temporal IoU is computed between the predicted interval and the annotated minimal interval. Spatial ground truth is a single box on the annotated evidence frame (Appendix~\ref{app:annotation}); spatial IoU is the IoU between that box and the predicted track's box on the same frame, and it is $0$ when the predicted interval does not contain that frame. A question is \emph{grounding-eligible} when it has an annotated interval, and \emph{box-annotated} when it additionally has an annotated box ($r^\star\neq\varnothing$). Temporal-only questions (no box, \eg, ``Did I lock the door?'') are excluded from all spatial metrics and from Acc@GQA rather than scored as $\mathrm{S\text{-}IoU}=0$; in Eq.~\ref{eq:referclaw-objective} they contribute $U_{\mathrm{grd}}=\mathrm{tIoU}$.

\paragraph{Populations and Definitions.} Tab.~\ref{tab:metric-populations} lists, for every reported metric, the fixed set of examples it is averaged over and what counts as a success. The 1,197 \emph{intent-ambiguous} and 5,043 \emph{well-specified} questions (the latter including the 201 history-resolved G4 questions) partition the 6,240 test inputs. \emph{Candidate-bearing inputs} are the 2,995 questions (48.0\%; 1,160 intent-ambiguous, 1,835 well-specified) for which the frozen Base IGA returns at least two candidates. This stratum is used only for evaluation, so that turns, Utility, and the entropy-bin analysis compare methods on the same inputs; every method constructs its own candidates at test time, and questions asked outside the stratum count toward the ask rate over all inputs. The 1,197 intent-ambiguous questions are the positives of Ask-P and Ask-R; the 201 G4 questions count toward the unnecessary-ask rate instead.

\paragraph{Runs and Aggregation.} Every number in a table comes from the same three runs (three fixed stream orders, identical seeds across methods), and the same checkpoint is used for all subsets of a row, so subset and overall values always refer to the same predictions; reported means are means of per-run values. GPU time includes candidate generation, response sampling, information-gain scoring, verification, and skill retrieval. Standard deviations and confidence intervals are in Appendix~\ref{app:ci}.

\FloatBarrier
\section{Additional Experimental Details}
\label{app:experimental-details}

\FloatBarrier
\subsection{Baselines and Fair Comparison}

We compare against complementary methods that isolate direct video answering, evidence retrieval, and active disambiguation. The \emph{direct-answer MLLM} is Qwen3.5-9B~\citep{qwen2026qwen35}. \emph{Grounded and streaming VideoQA} baselines include ReKV~\citep{di2025rekv}, RGA3~\citep{wang2025rga3}, and MyBuddy~\citep{qin2026buddyvqa}; these respectively test streaming memory retrieval, object-centric spatio-temporal referring, and ego-deictic reasoning with dialogue history. For joint answering and grounding (Tabs.~\ref{tab:egoblind-main} and~\ref{tab:grounded}), we further compare GeLM~\citep{chen2025gelm}, Open-o3-Video~\citep{meng2025openo3video}, GroundFormer~\citep{seo2026groundformer}, and EG-Reasoner~\citep{huang2026egvqa}, which return temporal or spatio-temporal evidence with each answer. \emph{Interactive clarification} baselines include the ask-capable ClearVQA model~\citep{jian2025clearvqa}, Plug-and-Play Clarifier~\citep{yang2026clarifier}, and Clarify-or-Answer with GRPO-CR~\citep{cao2026clarifyanswer}. Framework-level baselines that permit backbone substitution use Qwen3.5-9B, whereas architecture-specific methods use their official checkpoints. Every method receives the identical decoded frame stream, causal video prefix, and maximum visual-token budget, and the same access to the dialogue history; fixed-frame architectures uniformly subsample from that same prefix. The history a method sees at question $k$ of an episode consists of the earlier questions of that episode together with the method's \emph{own} earlier answers and clarification exchanges, so histories are identical across methods in access rights and in the user-side content but not in the assistant-side content; controls and ablations that must be compared at the same state are instead run as \emph{paired replays} on the logged entering state of the reference run (below). Gold ambiguity labels, resolved questions, spatial boxes, and future frames are hidden at prediction time. All interactive methods share the same respondent, maximum number of turns, and token and option budgets. We use released code and default prompts where available, changing only the input adapter required to accept a video clip instead of a single image. The \emph{asking-disabled} controls run IGA and EFA with the verifier but never invoke CCA; they are paired replays of the reference run (same adapter and skill version at each stream position, same entering history and seed), differing only in that the current question is never asked (Appendix~\ref{app:controls}).

\FloatBarrier
\subsection{Learning-from-Interaction Conditions}

Following MetaClaw's dynamic online evaluation~\citep{xia2026metaclaw}, we evaluate all adapting systems under identical inference and interaction budgets. For every ordered stream, interaction $k$ is first answered and scored by the current system, verified feedback (the annotated target intent, its evidence, and the judge's correctness verdict) is revealed only afterward, and any resulting skill or parameter update can influence interactions $k+1$ onward but never the current or preceding examples. The model, skill library, and optimization state are initialized identically at the start of each stream order and then carried forward without resets inside that stream. \textbf{\modelname~(Base)} runs the frozen pipeline. \textbf{\modelname~(Skills)} additionally enables fast adaptation: after each verified failure, the skill evolver synthesizes a short skill, and the three most relevant skills are retrieved by cosine similarity and inserted into the responsible agent's prompt. \textbf{\modelname~(Full)} combines skill adaptation with process-reward-guided policy optimization; every verified trajectory (successes and failures) enters the query buffer together with the skill version under which it was produced; skills take effect as soon as they are synthesized, so a buffer of 128 trajectories spans several skill generations. When the buffer is full and the server is not processing an online request, each failed step is re-executed with its own logged prompt, including the skills that were injected at the time, and only the LoRA adapter attributed to that step is updated with GRPO~\citep{shao2024deepseekmath}; the adapter version is constant within one consolidation window. The \emph{Reflexion-style memory} baseline stores a free-text reflection for each verified failure and retrieves the three most similar reflections into every agent prompt. The \emph{MetaClaw-style} baseline uses MetaClaw's generic skill schema (task, insight, action) without candidate topology or decisive contrast, and updates all three adapters with GRPO on the final-answer reward plus the same cost penalty; it receives the same update budget as \modelname~(Full).

\FloatBarrier
\subsection{Optimization and Hyperparameters}

\begin{table}[!htbp]
\centering
\footnotesize
\setlength{\tabcolsep}{3pt}\renewcommand{\arraystretch}{1.0}
\caption{\modelname~implementation and hyperparameters.}
\label{tab:referclaw-hparams}
\begin{tabular}{@{}p{0.17\linewidth}p{0.81\linewidth}@{}}
\toprule
Grounding & 1 fps causal stream; retrieved clips at 4 fps, $448\times448$; 8 temporal proposals; detection/text thresholds $0.25/0.20$ ($0.15/0.12$ for the single re-grounding pass) \\
Candidates & $N_{\max}=4$ after merging; $\kappa=0.5$; $\epsilon_0=0.2$; drop threshold $0.02$; question pool $|\mathcal C_t|\le6$ (3 multiple-choice, 3 open-ended) \\
Thresholds & $\tau_d=0.60$, $\tau_u=0.25$, $\tau_a=0.30$, $\tau_p=0.80$, $\tau_g=0.15$ \\
Budget & $T_{\max}=2$, $L_{\max}=24$ displayed tokens, $M_{\max}=4$ options incl.\ \emph{other/none} (one-shot stress test: $L_{\max}=64$, $M_{\max}=8$); generation cap 32 tokens per candidate question (longer candidates discarded) and 128 answer tokens \\
Respondent & 8 samples per $(c,z_i)$, smoothing $\delta=0.05$, free text mapped to classes by BGE-M3; $\lambda_g=0.50$, $\lambda_C=0.35$; $(w_E,w_T,w_F)=(0.45,0.25,0.30)$ human, $(0.55,0.15,0.30)$ simulator; components normalized to $[0,1]$ by validation 5th/95th percentiles \\
Learning & PRM weights $(\alpha,\beta,\lambda,\mu)=(0.50,0.50,0.35,1.00)$; one failure triggers skill synthesis ($\le6$ failures per call), BGE-M3 top-3 retrieval at cosine $\ge0.45$, $\le3$ injected skills; LoRA rank 16, $\alpha_{\mathrm{LoRA}}=32$, dropout $0.05$; query buffer 128; $G=8$; KL $0.02$ to Base; clip $0.2$; lr $5\times10^{-6}$, batch 64, one epoch per window; promotion probe 500+500 validation examples (target type $+1.0$, probe $\ge-0.5$, unnecessary-ask and cost $\le+0.01$) \\
Decoding & IGA/CCA temperature $0.2$, top-$p=0.9$; EFA/verifier temperature $0$ \\
\bottomrule
\end{tabular}
\end{table}

Videos are processed causally at 1 fps; the top retrieved intervals are resampled at 4 fps and $448\times448$ resolution for fine grounding. The three agent adapters and the PRM are trained in bfloat16 on eight NVIDIA H100 80GB GPUs and served with vLLM~\citep{kwon2023vllm}. AdamW uses $\beta_1=0.9$, $\beta_2=0.95$, weight decay $0.01$, gradient clipping at $1.0$, a 3\% linear warm-up, and cosine decay. Hyperparameters are selected once on the Augmented BuddyVQA validation split and then fixed for all test benchmarks. Cost weights $(w_E,w_T,w_F)$ and $\lambda_C$ are chosen by grid search over $\{0.1,\dots,0.6\}^3$ (normalized to sum to one) and $\{0,0.1,0.2,0.35,0.5,0.8\}$ to maximize validation utility (Eq.~\ref{eq:referclaw-objective}); the human and simulator cost weights are fitted separately and differ mainly in $w_T$ ($0.25$ for humans, $0.15$ for the simulator). The complete configuration is in Tab.~\ref{tab:referclaw-hparams}.

\FloatBarrier
\section{Additional Results}

\FloatBarrier
\subsection{Grounded Answering Results}
\label{app:extended-grounding}

\begin{table}[!htbp]
\centering
\small
\setlength{\tabcolsep}{4pt}
\caption{Grounded VQA (\%) on grounding-eligible questions of Augmented BuddyVQA and zero-shot Augmented EgoBlind-Intent. T/S: temporal/spatial. T-mIoU, T-R@.5, and Ans.\ are computed on all grounding-eligible questions, S-mIoU, S-R@.5, and Acc@GQA on the box-annotated subset (populations in Appendix~\ref{app:protocol}). Temporal-only baselines share our frozen Grounding DINO 1.5/SAM~2 spatial head. Superscripts on Ans.\ and Acc@GQA: change over the preceding row.}
\label{tab:grounded}
\begin{adjustbox}{max width=\textwidth}
\begin{tabular}{lrrrrrr|rrrrrr}
\toprule
& \multicolumn{6}{c|}{Augmented BuddyVQA} & \multicolumn{6}{c}{Augmented EgoBlind-Intent (zero-shot)} \\
Method & T-mIoU & T-R@.5 & S-mIoU & S-R@.5 & Ans. & GQA & T-mIoU & T-R@.5 & S-mIoU & S-R@.5 & Ans. & GQA \\
\midrule
GeLM~\citep{chen2025gelm} & 46.1 & 44.4 & 33.7 & 32.0 & 56.2 & 25.3 & 41.3 & 39.1 & 30.5 & 28.2 & 50.2 & 20.1 \\
RGA3~\citep{wang2025rga3} & 44.8 & 42.7 & 41.9 & 40.7 & 58.0 & 28.9 & 40.1 & 37.9 & 38.4 & 36.5 & 52.1 & 23.7 \\
Open-o3-Video~\citep{meng2025openo3video} & 49.3 & 47.2 & 45.5 & 43.9 & 60.1 & 33.4 & 44.7 & 42.4 & 41.2 & 39.5 & 54.0 & 27.0 \\
GroundFormer~\citep{seo2026groundformer} & 52.7 & 51.1 & 38.6 & 37.0 & 59.4 & 31.7 & 47.2 & 45.0 & 34.8 & 32.4 & 53.6 & 25.4 \\
EG-Reasoner~\citep{huang2026egvqa} & 51.6 & 49.9 & 43.2 & 41.7 & 61.0 & 35.0 & 46.5 & 44.1 & 39.8 & 37.7 & 54.8 & 27.6 \\
\midrule
\modelname~(Base, frozen) & 54.9 & 53.3 & 46.8 & 45.1 & 64.9\up{3.9} & 36.2\up{1.2} & 47.6 & 45.9 & 40.4 & 38.6 & 56.9\up{2.1} & 28.4\up{0.8} \\
\modelname~(Skills) & 56.6 & 55.1 & 48.5 & 47.0 & 68.6\up{3.7} & 38.9\up{2.7} & 48.9 & 47.2 & 41.6 & 39.8 & 58.8\up{1.9} & 30.1\up{1.7} \\
\modelname~(Full) & \textbf{58.3} & \textbf{57.0} & \textbf{50.3} & \textbf{48.7} & \textbf{72.9\up{4.3}} & \textbf{41.8\up{2.9}} & \textbf{50.1} & \textbf{48.4} & \textbf{42.6} & \textbf{40.9} & \textbf{60.7\up{1.9}} & \textbf{31.6\up{1.5}} \\
\bottomrule
\end{tabular}
\end{adjustbox}
\end{table}

\FloatBarrier
\subsection{Per-Category Transfer Breakdown}
\label{app:egoblind-breakdown}

On Augmented EgoBlind-Intent, \modelname~(Full) answer accuracy over all 2,036 instances (Tab.~\ref{tab:egoblind-main} uses the 1,447 grounding-eligible ones) is 66.8\% on instances without a blur or off-center flag ($n=1{,}291$) and 54.8\% on flagged instances ($n=745$), against 61.3\% and 43.6\% for EG-Reasoner; Acc@GQA on the box-annotated subset is 35.9\% \vers 24.7\% for the two groups. Clarification metrics transfer more completely than grounding: Ask-P and Ask-R are 82.6\% and 79.4\% on the 418 ambiguous instances (85.8\% and 83.1\% in-domain), and clarification success is 78.0\% (81.4\% in-domain).

\FloatBarrier
\subsection{Confidence Intervals}
\label{app:ci}

Standard deviations over the three stream orders are at most $0.6$ points for every accuracy in Tabs.~\ref{tab:buddy-main} and~\ref{tab:grounded}. 95\% paired-bootstrap confidence intervals resampled at the video level (10{,}000 resamples) are $[73.2,75.3]$ for \modelname~(Full), $[69.0,71.1]$ for \modelname~(Skills), $[65.1,67.4]$ for \modelname~(Base), $[57.3,59.7]$ for Clarify-or-Answer, $[58.0,60.4]$ for the MyBuddy reproduction, and $[50.2,52.6]$ for Direct Qwen3.5-9B (answer accuracy), and $[40.3,43.4]$, $[37.4,40.5]$, $[34.7,37.7]$, and $[33.5,36.5]$ for the Acc@GQA of \modelname~(Full), (Skills), (Base), and EG-Reasoner. Differences between \modelname~(Full) and every baseline, and between \modelname~(Base) and every frozen baseline, exclude zero. For ContextClarify, the 95\% interval of the enabled-minus-disabled difference on 42 items is $[+2.4,+21.4]$ points.

\FloatBarrier
\subsection{Which Design Helps Which Case}
\label{app:design-cases}

\begin{figure}[h]
    \centering
    \includegraphics[width=\linewidth]{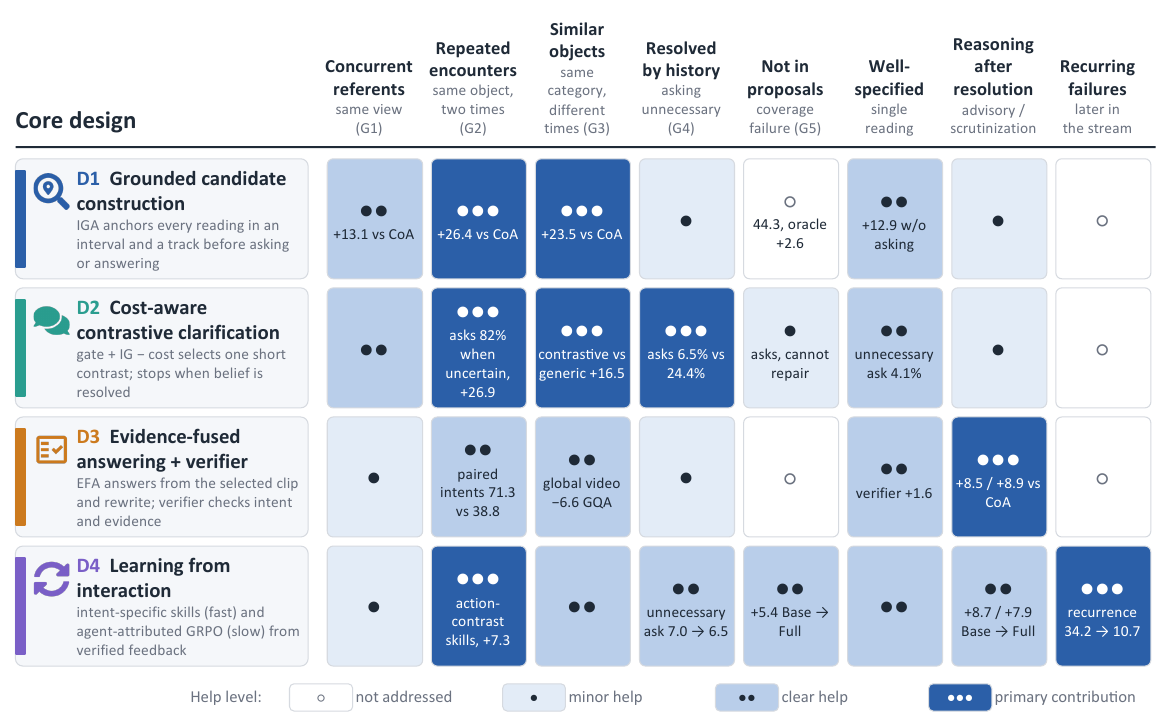}
    \caption{Which core design helps which case. Rows are the four core designs of \modelname; columns are the case types defined by the annotations (diagnostic groups G1--G5, well-specified questions, questions that require state or action reasoning after the referent is resolved, and failure types that recur along the interaction stream). Cell shading gives the qualitative level of help, and cell notes quote the result that supports it.}
    \label{fig:design-cases}
\end{figure}

Fig.~\ref{fig:design-cases} summarizes, per core design, the cases in which it makes a difference; the pattern is complementary rather than uniform. Grounded candidate construction is decisive where the competing readings are separated in time: because each candidate is anchored in its own interval and track, the two encounters of one object (G2) or two similar objects seen at different moments (G3) become distinct hypotheses that a question can name, which is where the advantage over the closest clarification baseline is largest (26.4 and 23.5 points against 13.1 when the referents are visible simultaneously, Fig.~\ref{fig:analysis-d}); removing the candidate set costs 11.3 Acc@GQA points and dropping the spatial tracks another 4.7 (Tab.~\ref{tab:agent-ablation}). The same design also carries most of the improvement on well-specified questions, where a single grounded candidate delivers selected evidence without any question (12.9 points over direct answering with asking disabled, Fig.~\ref{fig:clarification-benchmarks}), whereas it cannot help when the referent lies outside the proposals (G5), which remains the weakest group and whose headroom the coverage oracle bounds at 2.6 points (Tab.~\ref{tab:controls}).

Cost-aware contrastive clarification determines \emph{whether} and \emph{what} to ask. Its gate is what keeps the system silent when the history already resolves the reference (G4, asked on 6.5\% of inputs against 24.4\% for Clarify-or-Answer, with 11.4 points higher accuracy) and on well-specified questions (unnecessary-ask rate around 4\%), and its information-gain scoring concentrates questions where the candidate belief is uncertain (asked on 82.0\% of high-entropy inputs but 5.8\% of low-entropy ones, Fig.~\ref{fig:analysis-c}). What is asked matters as much as when: generic questions reduce clarification success from 81.4\% to 64.9\% (Tab.~\ref{tab:ablation}), and a heuristic that always asks about the top-two contrast recovers most of the accuracy only at higher cost (0.65 against 0.46 turns per candidate-bearing input, Tab.~\ref{tab:controls}). Clarification cannot repair a coverage failure, so on G5 it asks without benefit.

Evidence-fused answering acts after the intent is resolved and is therefore most visible on questions whose answer requires reasoning about state or action once the referent is known: Advisory and Scrutinization questions gain 8.5 and 8.9 points over Clarify-or-Answer (Tab.~\ref{tab:buddy-main}), replacing the selected clip by the global video costs 6.6 Acc@GQA points, and the paired-intent test, on which the same question must be answered differently for two intents, rises from 38.8\% to 71.3\% because the answer is produced from the selected evidence rather than from the most salient encounter. The verifier contributes a smaller but consistent 1.6 points by rejecting answers unsupported by the card.

Learning from interaction is the only design whose effect grows along the stream: recurrence of verified failure types falls from 34.2\% to 10.7\% and backward transfer is positive (4.1 points, Tab.~\ref{tab:evolution-analysis}), so its help concentrates on failure types that repeat. Its per-group effect is largest again on repeated encounters, driven by the action-contrast skills that account for 23\% of the learned CCA skills (Appendix~\ref{app:skill-example}), and it improves the Advisory and Scrutinization types by 8.7 and 7.9 points from Base to Full, whereas G5 improves only modestly and G4 mostly through fewer unnecessary questions (7.0\% to 6.5\%). Taken together, the four designs address different case types, which is why removing any one of them degrades a different column of Fig.~\ref{fig:design-cases} rather than uniformly lowering accuracy.

\FloatBarrier
\subsection{Controls that Isolate the Value of Clarification}
\label{app:controls}

\begin{table}[!htbp]
\centering
\small
\setlength{\tabcolsep}{5pt}
\caption{Controls that isolate the value of clarification (Augmented BuddyVQA, all test inputs). Turns: per candidate-bearing input; Cost: per asked input (``--'' when never asked). Top block: policies sharing the learned skills and adapters of \modelname~(Full); middle: the frozen Base pair and the online no-ask run; bottom: oracles using gold annotations at prediction time.}
\label{tab:controls}
\begin{adjustbox}{max width=\textwidth}
\begin{tabular}{lrrrrl}
\toprule
Variant & Ans. Acc. & Acc@GQA & Turns & Cost $\downarrow$ & What it measures \\
\midrule
\modelname~(Full) & 74.3 & 41.8 & 0.46 & 0.42 & -- \\
\modelname~(Full), asking disabled (state-matched replay) & 66.9 & 39.9 & 0.00 & -- & immediate effect of asking, learned states \\
Heuristic policy (always ask top-2 contrast when $g_t{=}1$) & 71.0 & 39.6 & 0.65 & 0.52 & value of IG$-$cost scoring \\
Single agent, same structured steps, one adapter & 71.6 & 39.9 & 0.50 & 0.44 & value of role separation \\
\midrule
\modelname~(Base) & 66.2 & 36.2 & 0.47 & 0.45 & -- \\
\modelname~(Base), asking disabled (state-matched replay) & 60.5 & 34.8 & 0.00 & -- & immediate effect of asking, frozen system \\
Online no-ask run (Base + both loops, CCA disabled throughout) & 64.8 & 38.3 & 0.00 & -- & learning without ever asking \\
\midrule
Oracle: gold intent inserted into candidate set & 76.9 & 44.6 & 0.41 & 0.40 & candidate coverage ceiling \\
Oracle: gold resolved question as input & 79.8 & 47.3 & 0.00 & -- & cost of intent resolution \\
\bottomrule
\end{tabular}
\end{adjustbox}
\end{table}

Both state-matched replays keep IGA, EFA, and the verifier but never invoke CCA: each question is re-run with the checkpoint, skill version, entering history, and seed that the reference run had at that stream position, so reference and replay differ only on inputs with $n_0\ge2$. Because the replayed states of \modelname~(Full) were themselves learned from earlier clarifications, the 7.4-point difference (74.3 \vers 66.9) is the \emph{immediate} effect of asking at matched states; the frozen pair gives 5.7 points. The \emph{online no-ask run} instead measures learning without any clarification: an independent prequential run from Base with both loops but CCA disabled reaches 64.8\%, 4.3 points above the frozen asking-disabled pipeline and 9.5 below \modelname~(Full), the latter being the total contribution of clarification including its effect on what is learned. The heuristic policy asks about the top-two candidates whenever the gate of Eq.~\ref{eq:clarification-gate} opens and therefore asks on every gate-open input (52.1\% of candidate-bearing inputs against 38.7\%); the single-agent control executes the same structured steps as one prompt sequence on one adapter. The oracle rows provide diagnostic headroom: inserting the gold intent adds 2.6 accuracy points and supplying the gold resolved question adds 5.5. These interventions do not provide an additive causal decomposition of the remaining errors.

\FloatBarrier
\subsection{Fine-Grained Ablation}
\label{app:fine-ablation}

\begin{table}[!htbp]
\centering
\small
\setlength{\tabcolsep}{5pt}
\caption{Fine-grained ablation on Augmented BuddyVQA; columns as in Tab.~\ref{tab:agent-ablation}. Within each block only the named component is changed.}
\label{tab:ablation}
\begin{adjustbox}{max width=\textwidth}
\begin{tabular}{lrrrrrrrrr}
\toprule
Variant & Ans. Acc. & Acc@GQA & Clar. Succ. & Ask (\%) & Cost (asked) $\downarrow$ & $\overline{U}_{\mathrm{ans}}$ & $\overline{U}_{\mathrm{grd}}$ & $\overline{\Sigma C}$ $\downarrow$ & Utility $\uparrow$ \\
\midrule
\modelname~(Full) & \textbf{74.3} & \textbf{41.8} & \textbf{81.4} & \textbf{38.7} & \textbf{0.42} & \textbf{0.732} & \textbf{0.328} & \textbf{0.163} & \textbf{0.733} \\
\midrule
\multicolumn{10}{l}{\emph{Cost-aware clarification}} \\
$-$ ambiguity gate & 72.5 & 40.4 & 73.6 & 70.0 & 0.69 & 0.710 & 0.320 & 0.483 & 0.387 \\
$-$ expected information gain (confidence-only trigger) & 71.9 & 40.0 & 70.1 & 42.0 & 0.47 & 0.708 & 0.316 & 0.197 & 0.669 \\
$-$ cost penalty ($\lambda_C=0$) & 73.8 & 41.4 & 82.0 & 52.1 & 0.76 & 0.732 & 0.326 & 0.396 & 0.499 \\
One-shot super-question (relaxed length budget) & 71.3 & 39.4 & 68.5 & 52.1 & 0.61 & 0.708 & 0.313 & 0.318 & 0.547 \\
Generic rather than contrastive clarification & 71.1 & 38.8 & 64.9 & 40.0 & 0.45 & 0.702 & 0.312 & 0.180 & 0.678 \\
\midrule
\multicolumn{10}{l}{\emph{Learning from interaction: fast loop}} \\
$-$ fast loop & 71.2 & 38.6 & 77.2 & 40.0 & 0.43 & 0.704 & 0.307 & 0.172 & 0.685 \\
Random skill retrieval & 72.0 & 39.4 & 78.0 & 40.0 & 0.44 & 0.715 & 0.315 & 0.176 & 0.696 \\
Top-1 rather than top-3 retrieval & 73.3 & 40.7 & 80.2 & 39.0 & 0.43 & 0.727 & 0.321 & 0.168 & 0.719 \\
Generic rather than intent-specific skill schema & 72.5 & 39.9 & 78.8 & 39.0 & 0.43 & 0.715 & 0.318 & 0.168 & 0.706 \\
$-$ support--query skill-generation separation & 72.2 & 39.5 & 78.5 & 39.0 & 0.43 & 0.709 & 0.318 & 0.168 & 0.700 \\
\midrule
\multicolumn{10}{l}{\emph{Learning from interaction: slow loop}} \\
$-$ slow loop (Skills only) & 70.1 & 38.9 & 79.4 & 40.0 & 0.43 & 0.691 & 0.312 & 0.172 & 0.675 \\
Final-answer-only reward & 72.6 & 39.4 & 79.1 & 43.0 & 0.47 & 0.716 & 0.310 & 0.202 & 0.669 \\
$-$ grounding reward & 73.1 & 37.6 & 80.3 & 39.0 & 0.43 & 0.722 & 0.303 & 0.168 & 0.706 \\
$-$ target-progress reward $\Delta_t^{\star}$ & 72.9 & 40.7 & 76.8 & 45.0 & 0.49 & 0.716 & 0.313 & 0.221 & 0.651 \\
Undirected entropy reduction $\Delta H_t$ instead of $\Delta_t^{\star}$ & 73.4 & 40.9 & 78.9 & 41.0 & 0.44 & 0.721 & 0.324 & 0.180 & 0.703 \\
Update all agent adapters & 72.3 & 39.4 & 78.3 & 42.0 & 0.46 & 0.713 & 0.312 & 0.193 & 0.676 \\
\bottomrule
\end{tabular}
\end{adjustbox}
\end{table}

\paragraph{Cost-Aware Clarification.} Removing the ambiguity gate makes the system ask on 70.0\% of candidate-bearing inputs instead of 38.7\%, raises the cost per asked input to 0.69, and reduces accuracy by 1.8 points because well-specified inputs are needlessly perturbed (lowest Utility, 0.387). Removing the cost penalty preserves most of the accuracy (73.8) but asks on every gate-open input (52.1\%, the same fraction as the one-shot policy and the no-stopping variant of Tab.~\ref{tab:agent-ablation}) at nearly twice the cost per asked input (0.76). The one-shot super-question, run with a relaxed length budget ($L_{\max}=64$, $M_{\max}=8$), loses 3.0 points despite a Cost of 0.61, and generic questions reduce clarification success from 81.4\% to 64.9\%.

\paragraph{Learning Loops.} Disabling the fast loop costs 3.1 accuracy points; random retrieval and a generic skill schema trail the full model by 2.3 and 1.8 points, so the gain comes from targeted intent skills rather than a longer prompt, and removing the support--query separation costs 2.1 points. The slow loop adds 4.2 points over Skills only: removing the grounding reward causes the largest Acc@GQA decrease (41.8 to 37.6), removing the target-progress term $\Delta_t^{\star}$ mainly reduces clarification success (81.4\% to 76.8\%), replacing it by an undirected entropy reduction costs 0.9 points and 2.5 points of clarification success, and updating every adapter is 2.0 points below agent-attributed updates.

\FloatBarrier
\subsection{Clarification Quality}
\label{app:clarification-quality}

\begin{table}[!htbp]
\centering
\small
\setlength{\tabcolsep}{5pt}
\caption{Clarification quality on Augmented BuddyVQA (\%). Amb-F1: F1 of the ambiguity flag; Ask-P/R: precision and recall of asking, with the 1,197 intent-ambiguous questions as positives; Clar.-Acc.: target intent selected after clarification, over the questions each method asked (asked) or over the 612 asked by all five methods (common); Intent-Acc.: final answer addresses the target intent, over all intent-ambiguous questions.}
\label{tab:clarification-analysis}
\begin{adjustbox}{max width=\textwidth}
\begin{tabular}{lrrrrrr}
\toprule
Method & Amb-F1 & Ask-P & Ask-R & Clar.-Acc. (asked) & Clar.-Acc. (common) & Intent-Acc. \\
\midrule
ClearVQA~\citep{jian2025clearvqa} & 72.6 & 68.8 & 76.2 & 66.1 & 64.1 & 58.3 \\
Plug-and-Play Clarifier~\citep{yang2026clarifier} & 75.4 & 72.0 & 76.8 & 69.5 & 67.8 & 61.0 \\
Clarify-or-Answer~\citep{cao2026clarifyanswer} & 78.1 & 76.3 & 79.6 & 73.7 & 71.9 & 64.9 \\
\modelname~(Base) & 81.0 & 82.4 & 80.9 & 77.5 & 76.3 & 72.8 \\
\modelname~(Full) & \textbf{84.3} & \textbf{85.8} & \textbf{83.1} & \textbf{81.4} & \textbf{80.2} & \textbf{77.6} \\
\bottomrule
\end{tabular}
\end{adjustbox}
\end{table}

In Tab.~\ref{tab:clarification-analysis}, \modelname~improves Amb-F1 by 6.2~points and ask precision by 9.5~over Clarify-or-Answer, so its gain is not obtained by asking more often. Because methods ask on different subsets, we also report clarification accuracy on the 612~ambiguous examples asked by all five methods; the ordering is unchanged and the gap widens slightly (80.2 \vers 71.9). Intent accuracy over all ambiguous examples, which penalizes both wrong resolutions and failures to ask, rises from 64.9\% to 77.6\%. Internally, the realized target progress $\mathrm{TP}=\sum_t\Delta_t^{\star}$ (the target-progress term of Eq.~\ref{eq:intent-prm}, in nats, computed on the fixed candidate scope of Appendix~\ref{app:reward-details}) averages $0.62$ per asked question for \modelname~(Full), and 92\% of asked questions with $\mathrm{TP}>0.5$ end with the target intent selected, confirming that belief moves toward the intended interpretation rather than toward a confidently wrong one.

\FloatBarrier
\subsection{Example Traces}
\label{app:traces}

\begin{figure}[t]
\centering
\includegraphics[width=\linewidth]{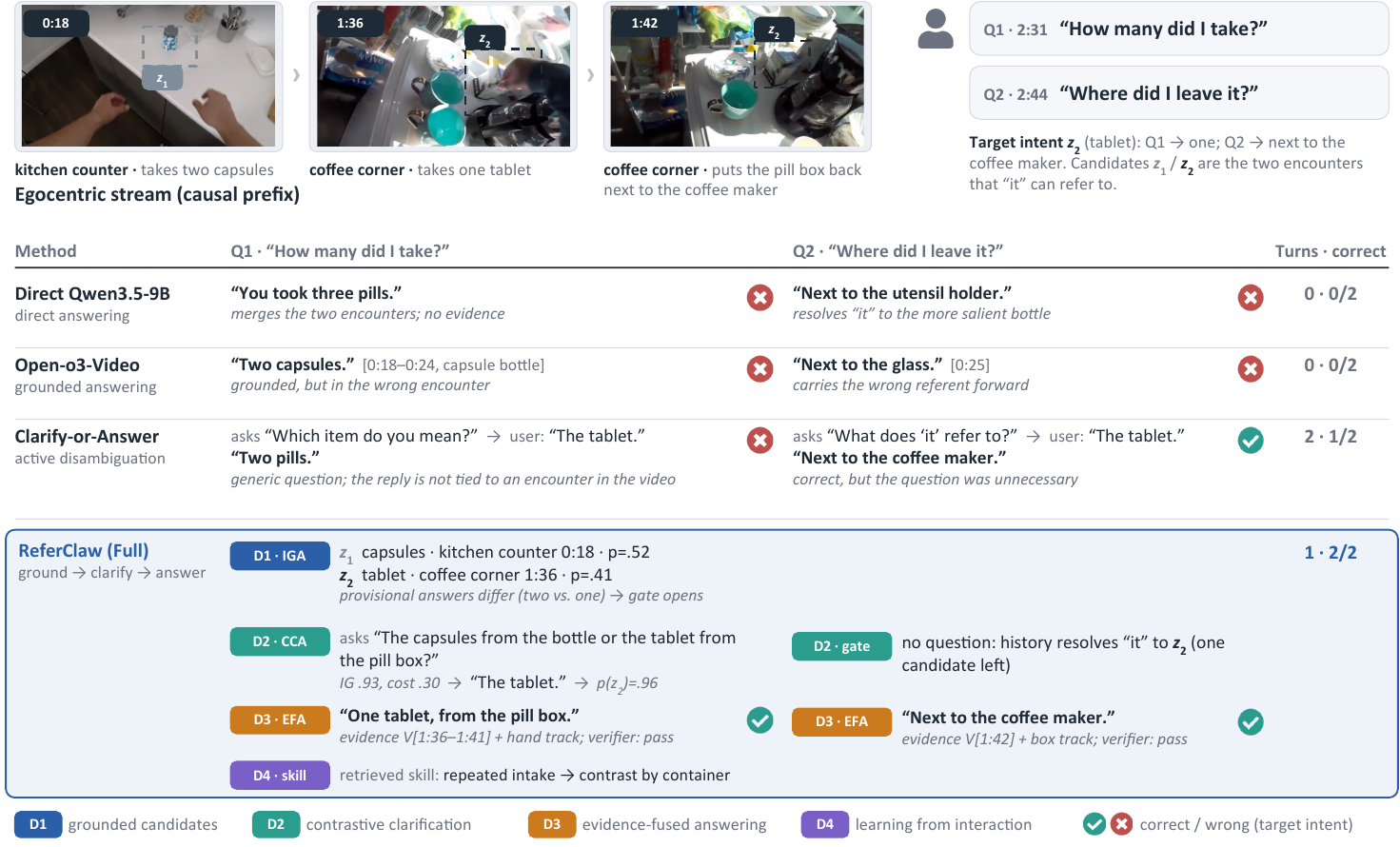}
\caption{Qualitative comparison on a two-question episode. $z_1$, $z_2$: grounded candidates (dashed boxes: their tracks); tags mark the core design behind each \modelname~step (D1--D4 as in Fig.~\ref{fig:design-cases}); \checkmark/$\times$: correct or wrong for the target intent $z_2$.}
\label{fig:qual-example}
\end{figure}

\paragraph{Qualitative Comparison.} Fig.~\ref{fig:qual-example} follows an episode in which the user takes two capsules from a bottle on the kitchen counter and, later, one tablet from a pill box at the coffee corner, and then asks ``How many did I take?'' and ``Where did I leave it?''. Direct answering merges the two encounters, and grounded answering localizes its evidence in the capsule encounter, an error that carries over to the follow-up. Clarify-or-Answer obtains the intended referent with a generic question, but because the reply is not tied to an encounter in the video it still answers from the more salient one, and it asks again although the dialogue already resolves ``it''. \modelname~grounds both encounters before asking (D1), so a single contrastive question separates them (D2), guided by a skill learned earlier in the stream that contrasts repeated intakes by container (D4); it then answers from the selected clip under verification (D3) and answers the follow-up without a question because the resolved referent leaves a single candidate.

\paragraph{Successful Trace.} Video prefix 1:50--2:30 (kitchen, then study). Question at 2:30: ``Is it still on?'' IGA: $d_0=0.81$; candidates $z_1$ (stove, 1:50--2:07, kitchen) with $p=0.41$ and $z_2$ (desk lamp, 2:20--2:30, study) with $p=0.52$, $p(z_\varnothing)=0.07$; provisional answers differ ($\bar a_1=$ no, $\bar a_2=$ yes; $\operatorname{Dis}=0.92$), $g_0=1$. CCA: pool of five questions; selected $c_0=$ ``The stove or the desk lamp?'' (options: stove, desk lamp, other/none), $\operatorname{IG}_0=0.88$ bits, $C_{\mathrm L}=0.28$, net utility $0.78$. Response: ``The lamp.'' Posterior: $p_1(z_2)=0.94$. Controller: $\max p_1\ge\tau_p$, \textsc{Answer}. EFA on $V[2{:}20,2{:}30]$ with the lamp track: ``Yes, the desk lamp is still on'' (\texttt{video}); verifier \emph{pass}. Correct.

\paragraph{Failure Trace.} Same episode; question at 2:32: ``Did I unplug it?'' after a dialogue turn about the toaster. IGA: candidates $z_1$ (toaster, 2:03) $p=0.60$ and $z_2$ (phone charger, 2:26) $p=0.32$; $d_0=0.72$, $\operatorname{Dis}=0.85$, $g_0=1$. CCA selected ``The toaster or the phone charger?'' Response: ``The toaster.'' EFA answered ``Yes, you unplugged it'' from a frame at 2:03 in which the hand occludes the socket; the verifier passed because the claim was labeled \texttt{video} and the card described ``hand at the toaster plug''. Annotated answer: no. Attribution: EFA (target covered and grounded; answer unsupported). This trace belongs to the occlusion category of the failure audit and illustrates the verifier's limitation as a prompted consistency check.

\FloatBarrier
\subsection{Model Dynamics and Failure Modes}
\label{app:model-analysis}

\paragraph{The Value of Interaction Is Concentrated in Uncertain Cases.} Fig.~\ref{fig:analysis-c} bins the 2,995 candidate-bearing test questions by the initial entropy $H(p_0)$ of the frozen Base candidate set; the low-, medium-, and high-entropy bins contain 1,288, 1,168, and 539~questions (43.0\%, 39.0\%, 18.0\%). To attribute the gain to asking rather than to learned grounding, each bin compares \modelname~(Full) with its paired asking-disabled replay (Tab.~\ref{tab:controls}) and with direct answering. \modelname~asks on only 5.8\% of the low-entropy bin, where asking adds 2.0~points (72.5~\vers~70.5), whereas it asks on 55.0\% and 82.0\% of the medium- and high-entropy bins, where asking adds 24.9~and 26.9~points (75.9~\vers~51.0~and 68.8~\vers~41.9); the gains over direct answering are 14.5, 29.9, and 29.9~points. Because the replay fixes the entering history, the two branches differ only on these 2,995 inputs, over which asking adds 15.4 points, corresponding to the 7.4-point Overall difference of Tab.~\ref{tab:controls}. This monotonic pattern explains why adaptive stopping dominates a uniform interaction depth: most inputs need little assistance, but the difficult tail benefits strongly from a well-chosen question.

\paragraph{Residual Failures Point to Perception Rather than to the Clarification Loop.} A manual audit of 500 incorrect \modelname~(Full) predictions attributes 31\% to severe occlusion or motion blur, 27\% to multiple visually near-identical referents, 23\% to information absent from the observed prefix, 12\% to long-range identity drift, and 7\% to invalid or noisy user responses. The first and fourth categories are grounding failures, and the third cannot be repaired by a better question. Future work should combine stronger persistent tracking with requests for \emph{new visual evidence} (\eg, asking the user to look toward the shelf), which our formulation can express as an action with cost but which we have not evaluated.

\FloatBarrier
\subsection{Efficiency and Respondent Cost}
\label{app:efficiency}

\begin{table}[!htbp]
\centering
\small
\setlength{\tabcolsep}{4.5pt}
\caption{Accuracy--cost trade-off on Augmented BuddyVQA. Ask rate: all inputs / candidate-bearing inputs; Outside: inputs outside the candidate-bearing stratum on which the method asked; Turns: per input over all / candidate-bearing / asked inputs; Tok./q.: tokens per question; Human time: seconds per response, measured in the user study or predicted by the calibrated regression ($\ddagger$); GPU time: seconds per input; Cost: per asked input / per input. The one-shot super-question uses a relaxed length budget.}
\label{tab:efficiency-analysis}
\begin{adjustbox}{max width=\textwidth}
\begin{tabular}{lrrrrrrrrrr}
\toprule
Method & Acc. & Ask rate (\%) & Outside & Turns (all / cand. / asked) & Tok./q. & Human time (s) & GPU time (s) & Cost (asked) $\downarrow$ & Cost (all) $\downarrow$ \\
\midrule
Direct Qwen3.5-9B & 51.4 & 0.0 / 0.0 & 0 & 0.00 / 0.00 / -- & -- & -- & 3.6 & -- & 0.000 \\
ClearVQA~\citep{jian2025clearvqa} & 55.7 & 21.2 / 43.1 & 35 & 0.28 / 0.57 / 1.32 & 23.3 & 6.0$^\ddagger$ & 5.0 & 0.59 & 0.125 \\
Plug-and-Play Clarifier~\citep{yang2026clarifier} & 57.4 & 20.4 / 41.7 & 28 & 0.28 / 0.58 / 1.38 & 22.2 & 5.9$^\ddagger$ & 5.4 & 0.63 & 0.129 \\
Clarify-or-Answer~\citep{cao2026clarifyanswer} & 58.5 & 20.0 / 41.0 & 20 & 0.26 / 0.53 / 1.28 & 24.0 & 6.9 & 5.1 & 0.55 & 0.110 \\
Fixed two-turn clarification & 68.8 & 25.0 / 52.1 & 0 & 0.50 / 1.04 / 2.00 & 18.4 & 4.9$^\ddagger$ & 7.8 & 0.86 & 0.215 \\
One-shot super-question (relaxed budget) & 71.3 & 25.0 / 52.1 & 0 & 0.25 / 0.52 / 1.00 & 47.5 & 11.1$^\ddagger$ & 5.8 & 0.61 & 0.153 \\
\modelname~(Base) & 66.2 & 18.8 / 39.2 & 0 & 0.23 / 0.47 / 1.21 & 21.7 & 5.3$^\ddagger$ & 4.9 & 0.45 & 0.085 \\
\modelname~(Full) & \textbf{74.3} & \textbf{18.6 / 38.7} & 0 & \textbf{0.22 / 0.46 / 1.19} & \textbf{20.6} & 5.1 & 5.0 & \textbf{0.42} & \textbf{0.078} \\
\bottomrule
\end{tabular}
\end{adjustbox}
\end{table}

Tab.~\ref{tab:efficiency-analysis} gives the operating points of Fig.~\ref{fig:analysis-a}, whose horizontal axis is the cost per input over all inputs. The fixed two-turn and one-shot policies ask on every gate-open input (25.0\% of all inputs), so their turn counts bound what a gate-triggered policy can spend. \modelname~(Full) reaches the highest accuracy while asking on the fewest inputs among the interactive methods, with the fewest questions per asked input and the shortest questions, at a GPU time (5.0\,s) within the range of the other interactive systems. Because frozen baselines and the adapting \modelname~(Full) are different regimes, the figure also marks the frozen \modelname~(Base), which dominates every frozen baseline on both axes (66.2\% at 0.085 against 58.5\% at 0.110 for Clarify-or-Answer).

\FloatBarrier
\subsection{Human Respondents}
\label{app:human}

\begin{table}[!htbp]
\centering
\small
\caption{Human-respondent evaluation on the 150 held-out examples of the user study (10 participants). Confusing: fraction of questions rated unanswerable or confusing. Sim.\ agreement: fraction of interactions in which the simulator's response falls in the same class as the human's. The last two columns give simulator accuracy on the same examples.}
\label{tab:human}
\begin{adjustbox}{max width=\textwidth}
\begin{tabular}{lrrrrrrr}
\toprule
Method & Intent resolved (\%) & Ans. Acc. (\%) & Resp. time (s) & Confusing (\%) & Sim. agreement (\%) & Acc. Qwen3-32B & Acc. Llama-3.3-70B \\
\midrule
Clarify-or-Answer & 70.0 & 57.3 & 6.9 & 12.7 & 85.9 & 58.1 & 57.6 \\
\modelname~(Full) & \textbf{80.0} & \textbf{72.0} & \textbf{5.1} & \textbf{4.3} & 88.6 & 73.7 & 73.4 \\
\bottomrule
\end{tabular}
\end{adjustbox}
\end{table}

Tab.~\ref{tab:human} makes the human study visible on the held-out half, which was not used to fit the response model or the response-time regression. With human respondents, \modelname~resolves the target intent on 80.0\% of examples and answers 72.0\% correctly, against 70.0\% and 57.3\% for Clarify-or-Answer, while requiring 1.8 fewer seconds per response and producing a third as many questions rated unanswerable or confusing. Simulator accuracy is 1.7 points higher than human accuracy for \modelname~and 0.8 for the baseline, so the simulator is slightly optimistic for both, and response-class agreement between simulator and humans is 86--89\%. Replacing the evaluation simulator with Llama-3.3-70B, which was not used for any design decision, changes accuracy by at most 0.5 points for either method, suggesting that the question-selection policy is not tuned to the particular simulator used during development.

\FloatBarrier
\subsection{Learning Dynamics}
\label{app:learning-dynamics}

\begin{table}[!htbp]
\centering
\small
\setlength{\tabcolsep}{4.5pt}
\caption{Learning dynamics on Augmented BuddyVQA. Q1--Q4: prequential accuracy on stream quartiles; Probe: accuracy on a fixed held-out probe of 1,000 examples after the stream; Recurrence: verified errors followed by the same failure type within 200 questions; BWT: accuracy change on first-quartile examples re-evaluated after the stream; Update: GPU-hours per 1,000 interactions.}
\label{tab:evolution-analysis}
\begin{adjustbox}{max width=\textwidth}
\begin{tabular}{lrrrrrrrr}
\toprule
Condition & Q1 & Q2 & Q3 & Q4 & Probe & Recurrence $\downarrow$ & BWT $\uparrow$ & Update GPU-h $\downarrow$ \\
\midrule
\modelname~(Base) & 66.0 & 66.1 & 66.3 & 66.4 & 65.8 & 34.2 & 0.0 & 0.0 \\
Base + Reflexion-style memory & 66.0 & 67.2 & 67.9 & 68.5 & 66.7 & 27.8 & $+$0.4 & 0.2 \\
Base + MetaClaw-style adaptation & 66.5 & 68.9 & 70.6 & 72.1 & 68.6 & 21.3 & $+$1.1 & 3.1 \\
\modelname~(Skills) & 66.2 & 69.3 & 71.5 & 73.3 & 70.0 & 18.5 & $+$1.9 & 0.3 \\
\modelname~(Full) & \textbf{66.5} & \textbf{72.6} & \textbf{77.0} & \textbf{81.2} & \textbf{75.2} & \textbf{10.7} & \textbf{$+$4.1} & \textbf{2.8} \\
\bottomrule
\end{tabular}
\end{adjustbox}
\end{table}

The quartile columns show prequential accuracy along the stream, while the probe column evaluates the final checkpoint on an unchanged held-out set so that the gain cannot reflect changing example difficulty. Skills improve the probe at 0.3 GPU-hours per 1,000 interactions because they take effect without weight updates; the slow loop's 2.8 GPU-hours are incurred asynchronously between online requests.

Fig.~\ref{fig:analysis-b} plots the probe accuracy of every checkpoint (after each 10\% of the stream, mean over the three orders; \modelname~(Base) stays at 65.8\%). \modelname~(Full) improves at every checkpoint, with the largest increments in the first half of the stream, where most first-of-type failures occur and the skill library grows fastest; the Skills-only curve flattens after 60\% of the stream, when new failures mostly repeat already-stored topologies, whereas the slow loop keeps consolidating gains until the end.

\FloatBarrier
\subsection{Coverage}
\label{app:coverage}

\modelname~(Full) returns an \textsc{Answer} on 96.1\% of Augmented BuddyVQA test inputs, a \textsc{Conditional} output on 2.3\%, and \textsc{Abstain} on 1.6\%; accuracy on the answered population is 77.3\%, and 61\% of \textsc{Conditional} outputs contain the correct answer paired with the target intent (counted as incorrect in all tables). Clarify-or-Answer answers 98.9\% of inputs with 59.1\% accuracy on the answered population; Direct Qwen3.5-9B answers 100\%. Reporting accuracy over all inputs, as in the main tables, therefore does not favor \modelname~through selective abstention.

\FloatBarrier
\section{Prompts and Templates}
\label{app:prompts-templates}

This section documents the prompts and structured interfaces used by \modelname. Each template specifies the model role, admissible inputs, decision constraints, and machine-readable output. Angle-bracketed strings denote runtime values. The same templates are used for human- and simulator-based evaluation; only the respondent model and cost calibration differ. GroundVQA, Grounding DINO 1.5 Pro, and SAM~2 produce temporal intervals, boxes, and tracks before prompting, so no LLM is allowed to invent spatial coordinates or unsupported evidence. Score normalization, Bayesian updates, expected information gain, cost aggregation, and controller actions are computed deterministically from the structured outputs.

\FloatBarrier
\subsection{Shared Interaction and Evidence Templates}

Every agent receives the original question, the dialogue history, the clarification history, retrieved active skills, and only the video evidence required by its role.

\textbf{Interaction Context Template.}
\begin{verbatim}
[QUESTION]
<original user question>

[DIALOGUE_HISTORY]
[ {"question": "<earlier user question>",
    "answer": "<assistant answer>"}, ... ]

[CLARIFICATION_HISTORY]
[ {"turn": 0, "question": "<c_0>", "response": "<y_0>"}, ... ]

[ACTIVE_SKILLS]
<zero to three skills retrieved for the current agent>
\end{verbatim}

\textbf{Evidence Card Template.}
\begin{verbatim}
{
  "evidence_id": "e1",
  "interval_sec": [<start>, <end>],
  "track_id": "track_1",            // null when non-localizable
  "boxes": {"<frame>": [x1, y1, x2, y2], ...},
  "representative_frames": [<frame ids>],
  "visual_description": "<literal visible content>",
  "grounding_score": <0.0--1.0>
}
\end{verbatim}

The visual description is restricted to observable content within the interval and track. It must not identify the intended candidate, infer the user's preference, or reveal a reference answer.

\FloatBarrier
\subsection{Intention Grounding Agent}

\textbf{IGA System Prompt.}
\begin{verbatim}
You are the Intention Grounding Agent for an egocentric video
assistant. Determine whether the user's intended question is
uniquely specified by the question, history, and supplied evidence.

Rules:
1. A deictic word alone does not imply ambiguity.
2. Mark ambiguity only if at least two visually supported
   interpretations remain plausible and can change the answer.
3. Create at most N_MAX candidates. Each candidate must contain
   a fully specified question and exactly one supplied evidence ID.
4. Preserve the original answer type. Do not reveal or insert the
   answer into a resolved question.
5. Use only supplied evidence cards. Never invent timestamps,
   regions, tracks, people, objects, actions, or user preferences.
6. Merge candidates that refer to the same entity and event.
7. If no card supports the question, return no_supported_candidate.
8. Return JSON only. Do not include chain-of-thought.

Inputs:
QUESTION: <question>
DIALOGUE_HISTORY: <dialogue_json>
CLARIFICATION_HISTORY: <history_json>
EVIDENCE_CARDS: <evidence_cards_json>
N_MAX: 4

Output schema:
{
  "ambiguity_score": <0.0--1.0>,
  "ambiguous_span": "<text or null>",
  "status": "unambiguous|ambiguous|no_supported_candidate",
  "candidates": [
    {
      "candidate_id": "z1",
      "resolved_question": "<fully specified question>",
      "evidence_id": "e1",
      "provisional_answer": "<short answer or unknown>",
      "support_score": <0.0--1.0>
    }
  ]
}
\end{verbatim}

Support scores are normalized outside the prompt with Eq.~\ref{eq:intent-posterior}; answer disagreement is computed deterministically from the provisional answers.

\FloatBarrier
\subsection{Clarification-Conditioned IGA Update}

After a user response, IGA narrows the existing candidate scope and requests targeted re-grounding only when the response selects \emph{other/none} or introduces a constraint not represented by the retained cards.

\textbf{IGA Update Prompt.}
\begin{verbatim}
You are updating an existing set of grounded intent candidates after
one clarification response. Do not restart from the full video.

Rules:
1. Interpret the response only with respect to the asked question.
2. Retain candidates consistent with the response and reject only
   candidates contradicted by it.
3. Do not create a new candidate unless the response selects
   other/none or supplies a missing observable constraint.
4. Request targeted re-grounding only for that new constraint.
5. Do not infer facts not stated by the user or visible in evidence.
6. Return JSON only.

Inputs:
ORIGINAL_QUESTION: <question>
CANDIDATES: <candidates_with_posterior_json>
EVIDENCE_CARDS: <evidence_cards_json>
CLARIFICATION_QUESTION: <c_t>
USER_RESPONSE: <y_t>

Output schema:
{
  "retained_candidate_ids": ["z1", "..."],
  "response_class": "<partition cell|other_none|uncertain|invalid>",
  "response_constraints": ["<atomic constraint>", "..."],
  "needs_targeted_regrounding": <true|false>,
  "regrounding_query": "<query or null>"
}
\end{verbatim}

The response likelihood and the Bayesian update in Eq.~\ref{eq:posterior-update} remain external to this prompt.

\FloatBarrier
\subsection{Cost-Aware Clarification Agent}

CCA proposes a small question pool; Eq.~\ref{eq:simple-cost} scores every valid proposal by $\operatorname{IG}_t(c)-\lambda_C C_x(c)$.

\textbf{CCA Question-Generation Prompt.}
\begin{verbatim}
You are the Cost-Aware Clarification Agent. Generate short questions
that distinguish the currently plausible grounded user intentions.

Rules:
1. Ask about exactly one observable contrast: identity, occurrence,
   location, or action.
2. Every contrast must be supported by the supplied evidence cards.
3. Prefer multiple choice when a few exclusive candidates cover most
   posterior mass; otherwise ask one short open-ended question.
4. Always include "Other/none".
5. Do not ask for the task answer, mention posterior scores, combine
   multiple attributes, repeat history, or ask a super-question.
6. Respect L_MAX tokens and M_MAX options.
7. Generate at most six candidates: up to three multiple-choice and
   three open-ended questions.
8. Return JSON only. Do not rank by fluency or provide reasoning.

Inputs:
QUESTION: <question>
HISTORY: <history_json>
CANDIDATES: <candidate_id, resolved_question, posterior>
EVIDENCE_CARDS: <evidence_cards_json>
L_MAX: 24
M_MAX: 4

Output schema:
{
  "questions": [
    {
      "question_id": "c1",
      "type": "multiple_choice|open_ended",
      "text": "<clarification question>",
      "options": ["<option>", "Other/none"],
      "contrast_type": "identity|occurrence|location|action",
      "candidate_partition": [["z1"], ["z2"]],
      "supporting_evidence_ids": ["e1", "e2"]
    }
  ]
}
\end{verbatim}

\FloatBarrier
\subsection{Predictive Respondent, Evaluation Simulator, and Human Interface}

The predictive respondent is prompted once per candidate $z_i$ to estimate $p_x(y\mid c,z_i,h_t)$ for CCA, and the evaluation simulator answers from the hidden target intent; what each role and each human participant receives, and never receives, is specified in Appendix~\ref{app:protocol}.

\textbf{Predictive Respondent Prompt (CCA-Internal).}
\begin{verbatim}
Answer the assistant's clarification question as a user whose
intended question is ASSUMED_INTENT. Reveal only what the
clarification asks. Give the shortest sufficient response, or
"uncertain" if ASSUMED_INTENT does not determine the answer.
Return JSON only: {"response": "<text>", "option": "<option|null>"}.

Inputs:
ORIGINAL_QUESTION: <question>
CLARIFICATION_HISTORY: <history_json>
CLARIFICATION_QUESTION: <question and options>
ASSUMED_INTENT: <resolved_question and evidence description of z_i>
\end{verbatim}

\textbf{Evaluation Simulator System Prompt.}
\begin{verbatim}
You simulate the user in an egocentric assistance interaction.
Answer the assistant's clarification from the hidden user intention.

Rules:
1. Respond as the user, not as an evaluator or assistant.
2. Reveal only the information requested by the clarification.
3. Do not reveal the task answer, candidate scores, or hidden fields.
4. For multiple choice, select the matching option; choose
   "Other/none" if no option matches.
5. For open-ended questions, give the shortest sufficient response.
6. Follow the respondent profile for uncertainty, invalid response,
   or no response. Do not become more cooperative than the profile.
7. Return JSON only.

Inputs:
ORIGINAL_QUESTION: <question>
CLARIFICATION_HISTORY: <history_json>
CLARIFICATION_QUESTION: <selected question and options>
HIDDEN_INTENT: <z_star without reference answer>
RESPONDENT_PROFILE: <human-calibrated or LLM-calibrated profile>

Output schema:
{
  "response": "<user response>",
  "valid": <true|false>,
  "response_status": "answered|uncertain|invalid|no_response"
}
\end{verbatim}

\FloatBarrier
\subsection{Evidence-Fused Answering Agent}
\label{app:efa-details}

\textbf{EFA System Prompt.}
\begin{verbatim}
You are the Evidence-Fused Answering Agent. Answer the resolved user
intention using only the selected video evidence and explicit user
clarifications.

Rules:
1. Answer the resolved question, not the ambiguous surface wording.
2. Treat USER_FACTS and VIDEO_FACTS as different sources.
3. Support every answer claim with the selected evidence card or an
   explicit user response, and label its source.
4. Do not use rejected candidates or unselected video intervals.
5. Do not invent object identity, timestamps, regions, or intent.
6. If INTENT_STATUS is "answer_determined_intent_unresolved", answer
   from the union of the supplied cards and say so in the packet.
7. Return concise JSON only. Do not include chain-of-thought.

Inputs:
ORIGINAL_QUESTION: <question>
RESOLVED_QUESTION: <q_tilde>
SELECTED_CANDIDATE: <candidate_id and posterior>
INTENT_STATUS: "resolved|answer_determined_intent_unresolved"
VIDEO_FACTS: <selected evidence card(s); frames carry the track box>
USER_FACTS: <clarification history>

Output schema:
{
  "answer": "<answer>",
  "resolved_question": "<q_tilde>",
  "evidence": {"evidence_id": "e1", "interval_sec": [<s>, <e>],
               "track_id": "track_1"},
  "confidence": <posterior of selected candidate>,
  "claim_sources": [
    {"claim": "<atomic claim>", "source": "video|user"} ]
}
\end{verbatim}

Conditional alternatives and abstention are decided by the controller (Tab.~\ref{tab:transitions}), not by EFA; each branch of a \textsc{Conditional} output is produced by a separate EFA call on its candidate and passes the verifier (\textsc{AnswerPath}).

\FloatBarrier
\subsection{Answer--Intent--Evidence Verifier}
\label{app:verifier}

\textbf{Verifier Prompt.}
\begin{verbatim}
Verify whether the proposed answer addresses the resolved intention
and whether each claim is supported by its declared source.

Rules:
1. Check answer--intent consistency and answer--evidence consistency.
2. A "video" claim must be visible in the supplied evidence card;
   a "user" claim must appear in USER_FACTS.
3. Do not repair an answer by inventing evidence or changing intent.
4. Choose revise only when unsupported text can be removed without
   changing the supported answer; otherwise choose abstain.
5. Return JSON only.

Inputs:
ORIGINAL_QUESTION: <question>
RESOLVED_QUESTION: <q_tilde>
EVIDENCE_CARD: <selected evidence card>
USER_FACTS: <clarification history>
ANSWER_PACKET: <EFA output JSON>

Output schema:
{
  "decision": "pass|revise|abstain",
  "intent_consistent": <true|false>,
  "evidence_consistent": <true|false>,
  "unsupported_claims": ["<claim>", "..."],
  "revised_answer": "<answer or null>"
}
\end{verbatim}

\FloatBarrier
\subsection{Controller Record}

The controller is non-generative; it applies Tab.~\ref{tab:transitions} to agent outputs and logs the following record for each turn.

\begin{verbatim}
{
  "turn": <t>,
  "ambiguity_score": <d_t>,
  "max_posterior": <max_i p_t(z_i)>,
  "uncovered_mass": <p_t(z_null)>,
  "clarification_gate": <0|1>,
  "selected_question_id": "<c_t or null>",
  "expected_information_gain_bits": <IG_t or null>,
  "respondent_cost": <C_x or null>,
  "net_utility": <IG_t - lambda_C C_x or null>,
  "response_class": "<cell|other_none|uncertain|invalid|null>",
  "action": "ASK|ANSWER|CONDITIONAL|ABSTAIN|REGROUND",
  "reason_code": "<deterministic rule identifier>"
}
\end{verbatim}

\FloatBarrier
\subsection{Process Review and Credit Assignment}

The process-review prompt is used only to attribute failures that the deterministic rules of Appendix~\ref{app:attribution} do not cover, and to audit training annotations after verified feedback is available; it is never exposed to the online agents. Target-progress terms and respondent costs are supplied as computed values.

\begin{verbatim}
You are annotating a completed trajectory after verified feedback.
Evaluate the disambiguation process, not only the answer.

Rules:
1. Check whether the intended candidate was covered and grounded.
2. Check whether the asked question separated the intended candidate
   from its strongest competitor.
3. Use the supplied target-progress terms and costs without changing them.
4. Flag missed ambiguity, unsupported candidates, repeated or
   multi-attribute questions, and incorrect stopping.
5. Attribute the primary failure to IGA, CCA, or EFA. Use none when
   the trajectory is valid or the evidence is genuinely unavailable.
6. Return JSON only and do not propose a new answer.

Inputs:
TRAJECTORY: <trajectory_json>
VERIFIED_INTENT: <z_star>
VERIFIED_EVIDENCE: <e_star>
VERIFIED_ANSWER: <correctness feedback>
TARGET_PROGRESS: <delta_star_by_turn>
RESPONDENT_COSTS: <cost_by_turn>

Output schema:
{
  "invalid_events": ["<event>", "..."],
  "responsible_agent": "IGA|CCA|EFA|none",
  "failure_type": "coverage|grounding|non_separating|unnecessary_ask|
                   inconsistent_answer|none",
  "short_rationale": "<one sentence>"
}
\end{verbatim}

\FloatBarrier
\subsection{Trajectory and Support--Query Records}

\begin{verbatim}
{
  "trajectory_id": "<id>",
  "skill_generation": <g>,
  "video_id": "<id>",
  "question": "<q>",
  "dialogue": <dialogue_json>,
  "history": <clarification_history_json>,
  "candidate_states": <candidates_and_posteriors_by_turn>,
  "evidence_cards": <evidence_cards_json>,
  "controller_actions": <controller_records_json>,
  "final_output": <EFA_packet_json or CONDITIONAL/ABSTAIN>,
  "verified_feedback": <feedback_json>,
  "process_components": {"r_final": .., "r_ground": ..,
      "sum_dH_plus": .., "sum_cost": .., "r_invalid": ..},
  "responsible_agent": "IGA|CCA|EFA|none",
  "split": "support_g|query_g_plus_1"
}
\end{verbatim}

Support and query records with the same failure event are never used in the same policy update.

\FloatBarrier
\subsection{Intent-Specific Skill Evolver and Injection}

\textbf{Skill Evolver Prompt.}
\begin{verbatim}
You are the skill evolver. Analyze verified failures and produce new
skills that prevent the same intent-disambiguation error without
increasing unnecessary clarification.

[VERIFIED_FAILURES]
<up to six support trajectories with failed step,
 failure type, and feedback>

[EXISTING_SKILLS]
<skill names, responsible agents, and triggers>

Instructions:
1. Generate one to MAX_NEW_SKILLS non-duplicate skills.
2. Each skill must target exactly one agent: IGA, CCA, or EFA.
3. Derive every visual pattern from observed evidence in the failures.
4. Store the ambiguity trigger, candidate topology, decisive visual
   contrast, corrected action, anti-pattern, and regression tests.
5. Do not store instance IDs, reference answers, posterior values,
   or advice to ask more questions by default.
6. A regression test must cover one ambiguous case and one
   well-specified case to protect the unnecessary-ask rate.
7. Return only a valid JSON array.

Output schema:
[
  {
    "name": "<lowercase-hyphenated-name>",
    "responsible_agent": "IGA|CCA|EFA",
    "trigger": "<deictic or elliptical failure condition>",
    "candidate_topology":
        "<e.g., same object, two encounters, same room, gap<60s>",
    "decisive_visual_contrast": "<identity/time/location/action>",
    "corrected_action": "<short executable instruction>",
    "anti_pattern": "<behavior to avoid>",
    "retrieval_key": "<trigger + topology + agent>",
    "regression_tests": [
      {"case": "ambiguous", "expected": "<behavior>"},
      {"case": "well_specified", "expected": "<behavior>"}
    ]
  }
]
\end{verbatim}

\textbf{Skill Injection Block.} At inference time, BGE-M3 retrieves at most three skills for the responsible agent; only the trigger, topology, corrected action, and anti-pattern are inserted, and regression tests remain in the store.
\begin{verbatim}
[ACTIVE_INTENT_SKILLS]
SKILL: <name>          AGENT: <IGA|CCA|EFA>
TRIGGER: <when this skill applies>
CANDIDATE_TOPOLOGY: <grounded configuration>
DECISIVE_CONTRAST: <identity|time|location|action>
ACTION: <corrected agent behavior>
ANTI_PATTERN: <behavior to avoid>
Apply a skill only when its trigger and topology match the current
input. The system prompt and interaction limits always take priority.
\end{verbatim}

\FloatBarrier
\subsection{Agent-Attributed Policy-Optimization Record}

The slow loop samples alternative outputs only from the adapter attributed by the failure attribution; other adapters and the visual backbone remain frozen. Each GRPO item uses the following record.
\begin{verbatim}
{
  "trajectory_id": "<id>",
  "skill_generation": <g_plus_1>,
  "responsible_agent": "IGA|CCA|EFA",
  "agent_input": <exact structured input at failed step>,
  "sampled_agent_outputs": [<G structured outputs>],
  "predicted_process_rewards": [<R_proc from the reward model>],
  "group_advantages": [<normalized advantages>],
  "active_skill_ids": ["<skill>", "..."]
}
\end{verbatim}

\end{document}